%% file: main_arxiv.tex
\pdfoutput=1
\documentclass[11pt]{article}
\usepackage[letterpaper,margin=1in]{geometry}
\usepackage[T1]{fontenc}
\usepackage{lmodern}
\usepackage{natbib}
\setcitestyle{authoryear,round,citesep={;},aysep={,},yysep={;}}

\usepackage{microtype}
\usepackage{amsmath,amssymb,amsthm,bm}
\usepackage{graphicx}
\usepackage{booktabs}
\usepackage{array}
\usepackage{xcolor}
\usepackage[colorlinks=true,linkcolor=blue!50!black,citecolor=blue!50!black,urlcolor=blue!50!black]{hyperref}
\usepackage{url}

\newtheorem{theorem}{Theorem}
\newtheorem{proposition}{Proposition}

\theoremstyle{definition}

\theoremstyle{remark}

\newcommand{\R}{\mathbb{R}}
\newcommand{\Srms}{\mathcal{S}}
\newcommand{\twoline}[2]{\begin{tabular}[b]{@{}c@{}}#1\\#2\end{tabular}}

\title{Rethinking Mamba-3 Integration Rules}

\author{%
  Ming-Ching Chang\textsuperscript{1}\quad
  Ting Yu Tsai\textsuperscript{1}\quad
  Davis Wertheimer\textsuperscript{2}\quad
  Felix X.-F. Ye\textsuperscript{3}\thanks{Corresponding author.} \quad An Yu \textsuperscript{1} \\[6pt]
  {\small\textsuperscript{1}Department of Computer Science, University at Albany, SUNY}\\
  {\small\textsuperscript{2}IBM T. J. Watson Research Center}\\
  {\small\textsuperscript{3}Department of Mathematics and Statistics, University at Albany, SUNY}\\[4pt]
  {\texttt{xye2@albany.edu}}}
\date{}

\begin{document}
\maketitle
\begingroup
\renewcommand{\thefootnote}{}
\NoHyper\footnotetext{We used generative AI tools, including Claude (Anthropic), Codex (OpenAI), and Gemini (Google), to discuss ideas, check mathematical derivations and proofs, write code, analyze results, prepare figures,  and edit the paper. We take full responsibility for all content.}\endNoHyper
\endgroup

\begin{abstract}
Mamba-3 learns a trapezoidal parameter $\lambda$ that weights the previous and current state-inputs of each step. Models trained with $\lambda$ fixed at $1/2$ reach nearly the same perplexity as models that learn it. Yet resetting a selected early layer to $\lambda=1/2$ costs 9--20 points of LAMBADA accuracy in each of four released SISO checkpoints. To find where $\lambda$ can matter, we study a single-head SISO surrogate with step sizes, state transitions and $\lambda$ frozen over depth. We prove that it converges to the same causal Volterra equation for every choice of $\lambda$ as the token grid becomes dense. The proof assumes that decay and rotation build up gradually and that inputs change smoothly between adjacent tokens. In the released checkpoints, early heads with memory horizons below two tokens break both assumptions. To locate the reset cost, we split a change of $\lambda$ exactly into a current-token change on the diagonal of the structured mask and a past-token change below it. The current-token change carries almost all of the one-layer reset cost, and most of this cost sits in heads with small mean $\lambda$ and short memory horizons. The past-token change has little effect in that layer but costs 13--15 points when applied in all layers. In 187M models trained from scratch, early heads with short memory horizons put most of their weight on the previous token and keep the current-token term small. Heads that learn the current-token weight drive it near zero, while heads with this weight fixed at $1/2$ compensate by making their $B$ and $C$ projections nearly orthogonal. This compensation explains why fixing $\lambda$ before training costs little. In heads that learn this weight, the two projections end up aligned, so only the small $\lambda$ keeps the current token out. Resetting $\lambda$ to $1/2$ afterward lets it back in.
\end{abstract}

\section{Introduction}\label{sec:intro}

\begin{figure}[t]
\centering
\includegraphics[width=\linewidth]{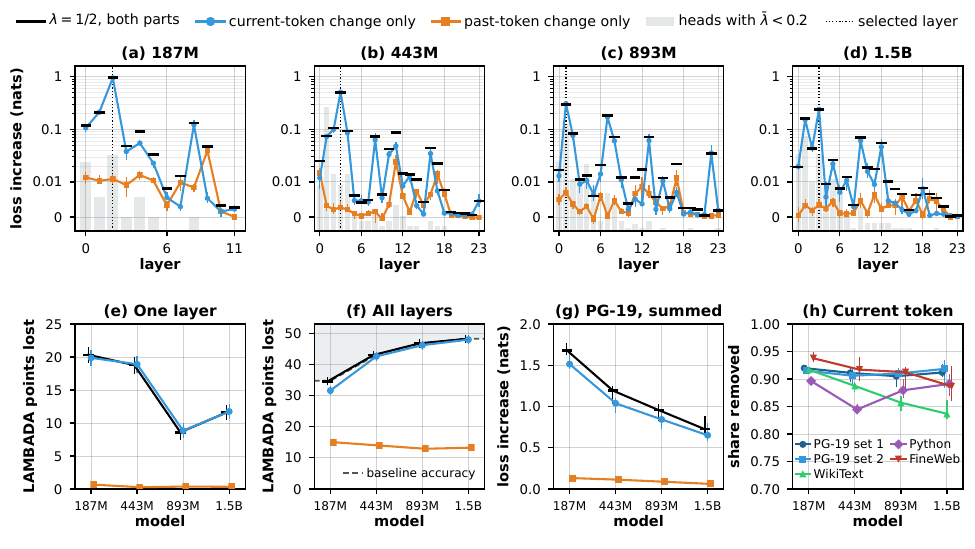}
\caption{The cost of resetting one layer to $\lambda=1/2$ comes mostly from the current-token change. Blue and orange lines show the current-token and past-token changes applied alone. (a--d) PG-19 loss increase per reset layer (linear below 0.01 nats). Grey bars show the fraction of heads with mean $\lambda<0.2$ (0--1 over the panel height), and dotted lines mark the layer selected for (e). (e,f) Mean LAMBADA accuracy lost for that layer and for all layers together (dashed in f: baseline). (g) Sum of the one-layer PG-19 increases. (h) Share of the summed increase recovered by restoring the current-token term, on PG-19 sets 1 and 2, WikiText-103, Python and FineWeb-Edu. Error bars are 95\% bootstrap intervals for the median, or the mean in (e,f).}
\label{fig:headline}
\vspace{-0.2in}
\end{figure}

Selective state space models (SSMs) carry information across tokens through a recurrent state \citep{gu2022s4,gu2023mamba,dao2024mamba2}. Mamba-3 weights the state-input of the current token by a learned trapezoidal parameter $\lambda$ and that of the previous token by $1-\lambda$, so $\lambda=1/2$ gives both equal weight. The original 440M-parameter study reports perplexities of 15.72 with learned weights and 15.76 with equal weights \citep[Table~8]{lahoti2026mamba3}. Yet in each of four released single-input, single-output (SISO) checkpoints, resetting a single early layer to equal weights lowers LAMBADA final-word accuracy \citep{paperno2016lambada} by 9--20 percentage points (Figure~\ref{fig:headline}e). Why do equal weights work well when fixed from the start, yet disrupt prediction when imposed afterward?

Mamba-3 motivates $\lambda$ through numerical integration, so we first ask where $\lambda$ can matter at all. Section~\ref{sec:refinement} studies a single-head SISO surrogate with step sizes, state transitions and $\lambda$ frozen over depth. As the token grid becomes dense, the surrogate converges to the same causal Volterra equation for every choice of $\lambda$ (Theorem~\ref{thm:volterra}). The proof assumes that decay and rotation build up gradually along the sequence (A1) and that inputs change smoothly between adjacent tokens (A2). In the surrogate, $\lambda$ can therefore matter only where one of these assumptions fails. In the released checkpoints, early short-horizon heads (memory horizons below two tokens) break (A1), and input jumps between adjacent tokens break (A2) in every head (Section~\ref{sec:native}).

To locate the cost of resetting $\lambda$ to $1/2$, we split any change of $\lambda$ exactly into a current-token change on the diagonal of the structured mask and a past-token change below it (Section~\ref{sec:setup}). The $B$ projection maps a token into the state, and the $C$ projection maps the state into the SSM output. The current-token term is the part of the SSM output that comes from the current token. It is proportional to $\lambda$ and to the inner product of the $B$ and $C$ projections of that token. In all four checkpoints (187M--1.5B), the change of this term carries almost all of the one-layer reset cost and reproduces the LAMBADA drop (Figure~\ref{fig:headline}e). The past-token change has little effect in that layer but costs 13--15 LAMBADA points when applied across all layers at once (Figure~\ref{fig:headline}f). Most of the current-token cost comes from a small group of heads with mean $\lambda$ below $0.2$. These heads mostly have short memory horizons. A head with constant parameters explains why: at a short memory horizon, past tokens carry little total weight, so raising a small $\lambda$ to $1/2$ makes a large change in the SSM output.

Training from scratch shows why short-horizon heads keep their current-token term small. We train 187M models on FineWeb-Edu \citep{penedo2024fineweb} under four rules that learn or fix the current- and past-token weights (Section~\ref{sec:training}), and fixing them barely changes the validation loss. Under every rule, early short-horizon heads put most of their weight on the previous token, a pattern common in sequence models \citep{fu2023h3,elhage2021framework,wang2025universality}. This pattern requires a small current-token term, which a head can obtain through $\lambda$ or through the inner product of its $B$ and $C$ projections. Heads that learn the current-token weight drive it near zero, while heads with this weight fixed at $1/2$ compensate by making the $B$ and $C$ projections of each token nearly orthogonal. In heads that learn this weight, the $B$ and $C$ projections end up aligned, so only the small $\lambda$ keeps the current token out. Resetting $\lambda$ to $1/2$ afterward therefore lets the current token back in.

\section{The rule in the structured mask}\label{sec:setup}

This section shows that $\lambda$ enters the SSM output only through the weights of the structured mask. The weight on the current token multiplies the inner product of the $B$ and $C$ projections, so a head can keep the current token out through either factor.

A SISO head of a Mamba-3 layer has head dimension $P$ and state size $N$. It runs $P$ copies of a SISO recurrence, one per coordinate of its SSM input $\bm x_i\in\R^P$. The copies share the $B$ and $C$ projections $B_i,C_i\in\R^N$ and the same state transitions. At token $i=1,\dots,T$, the layer computes $\bm x_i$, $B_i$, $C_i$, a step size $\Delta_i>0$, a scalar decay $A_i<0$, the trapezoidal parameter $\lambda_i\in[0,1]$ and rotation angles $\bm\omega_i\in\R^{N/2}$. It forms the state-input $\bm v_i=B_i\bm x_i^\top\in\R^{N\times P}$. The state $\bm h_i\in\R^{N\times P}$ and the SSM output $\bm y_i\in\R^P$ follow the exponential-trapezoidal recurrence
\begin{equation}\label{eq:recurrence}
\bm h_i=\alpha_i\mathbf R_i\bm h_{i-1}+\beta_i\mathbf R_i\bm v_{i-1}+\gamma_i\bm v_i,
\qquad
\bm y_i=\bm h_i^\top C_i,
\end{equation}
with $\bm h_0=0$, $\bm v_0=0$, $\beta_i=(1-\lambda_i)\Delta_i\alpha_i$ and $\gamma_i=\lambda_i\Delta_i$ \citep{lahoti2026mamba3}. Here $q_i=-\Delta_iA_i>0$ is the per-step decay and $\alpha_i=e^{\Delta_iA_i}=e^{-q_i}$. The rotation matrix $\mathbf R_i=\mathbf R(\bm\omega_i)$ (Mamba-3's data-dependent RoPE) rotates the $N/2$ coordinate planes of the state, so $\alpha_i\mathbf R_i$ is the one-step state transition. At the recurrence level, setting $\lambda\equiv1$ and $\bm\omega\equiv0$ gives the exponential-Euler form used by Mamba-2 \citep{dao2024mamba2}, while $\lambda\equiv1/2$ gives the two endpoints of each step equal weight, as in the classical trapezoidal rule. Each choice of $\lambda$ defines an integration rule for the state-input. Mamba-3 notes that $\alpha_i$ controls the memory horizon of each SSM \citep{lahoti2026mamba3}. Under a constant per-step decay $q$, the weight of a past state-input falls by a factor $e$ every $1/q$ tokens, so we call $1/q_i$ the memory horizon at token $i$. The released Mamba-3 layer adds a skip term $D\bm x_i$, with one scalar skip parameter $D$ per head, and multiplies $\bm y_i+D\bm x_i$ by a gate. Appendix~\ref{app:setup} gives the full layer as released. At a fixed layer input, any two rules share both the skip term and the gate, so their outputs differ only through the gated change in $\bm y_i$.

The SSM output $\bm y_i$ collects the state-inputs of all tokens up to $i$, each carried by the state transitions and weighted by the rule. Let $U_{ij}=(\alpha_i\mathbf R_i)\cdots(\alpha_{j+1}\mathbf R_{j+1})$ be the state transition from token $j<i$ to token $i$, with $U_{ii}=I$. Because the plane rotations commute, $U_{ij}=e^{-(H_i-H_j)}\mathbf R(\bm\Phi_i-\bm\Phi_j)$ with cumulative decay $H_i=\sum_{k\le i}q_k$ and cumulative angles $\bm\Phi_i=\sum_{k\le i}\bm\omega_k$. We write $F_{ij}=(U_{ij}\bm v_j)^\top C_i=(C_i^\top U_{ij}B_j)\bm x_j\in\R^P$ for token $j$'s state-input carried to token $i$ and multiplied by $C_i$, with $F_{i0}=0$. Unrolling \eqref{eq:recurrence} gives
\begin{equation}\label{eq:quadrature}
\bm y_i=\sum_{k=1}^{i}\Delta_k\bigl[(1-\lambda_k)F_{i,k-1}+\lambda_kF_{ik}\bigr],
\end{equation}
where step $k$ joins tokens $k-1$ and $k$ with weights $1-\lambda_k$ and $\lambda_k$ on its two endpoints. Grouping the sum by source token $j$ yields $\bm y_i=\sum_{j\le i}w_{ij}F_{ij}$ with lower-triangular mask weights
\[
(w_{ij})=\begin{pmatrix}
\gamma_1&0&0&\cdots\\
\eta_1&\gamma_2&0&\cdots\\
\eta_1&\eta_2&\gamma_3&\cdots\\
\vdots&\vdots&\vdots&\ddots
\end{pmatrix},\qquad
\eta_j=\lambda_j\Delta_j+(1-\lambda_{j+1})\Delta_{j+1},
\]
which form the size-two data-dependent convolution on the state-input in Mamba-3's structured mask \citep{lahoti2026mamba3}. The diagonal entries $\gamma_i$ weight the current token ($j=i$), while the off-diagonal entries $\eta_j$ weight past tokens ($j<i$).

The diagonal gives the current-token term of the SSM output,
\begin{equation}\label{eq:current}
\gamma_iF_{ii}=\lambda_i\Delta_i\,(B_i^\top C_i)\,\bm x_i.
\end{equation}
Since $\lambda_i$ and $B_i^\top C_i$ enter as a product, the term vanishes when $\lambda_i=0$ or when $B_i$ and $C_i$ are orthogonal. In either case, the direct contribution $(D+\gamma_iB_i^\top C_i)\bm x_i$ of the current token reduces to the skip term $D\bm x_i$ before the gate, and $\bm v_i$ can still enter the state at step $i+1$. Orthogonality of $B_i$ and $C_i$ does not constrain the past-token terms $F_{ij}=(C_i^\top U_{ij}B_j)\bm x_j$ with $j<i$. A head with $\lambda$ fixed at $1/2$ can therefore keep the current token out of its SSM output while still carrying past tokens, and Section~\ref{sec:training} shows that trained heads do so.

At a fixed layer input, two rules $\lambda$ and $\mu$ share $\Delta$, $\bm x$, $B$, $C$ and the state transitions. They differ only in the mask weights $w_{ij}$. By \eqref{eq:quadrature}, their outputs differ by
\begin{equation}\label{eq:difference}
\bm y_i^{\lambda}-\bm y_i^{\mu}=\sum_{k=1}^{i}\Delta_k(\lambda_k-\mu_k)\bigl(F_{ik}-F_{i,k-1}\bigr).
\end{equation}
Setting $a_k=\Delta_k(\lambda_k-\mu_k)$ for the change in $\gamma_k$ and grouping by source token $j$ yields
\begin{equation}\label{eq:presentpast}
\bm y_i^{\lambda}-\bm y_i^{\mu}
=\underbrace{a_i\,F_{ii}}_{\substack{\text{current-token}\\\text{change }\mathcal D_i}}
+\underbrace{\sum_{j<i}\,(a_j-a_{j+1})\,F_{ij}}_{\substack{\text{past-token}\\\text{change }\mathcal O_i}}.
\end{equation}
The current-token change $\mathcal D_i$ rescales the current-token term, while the past-token change $\mathcal O_i$ redistributes the contributions of earlier tokens. Because the past-token weight is $\eta_j=\lambda_j\Delta_j+(1-\lambda_{j+1})\Delta_{j+1}$, its change $a_j-a_{j+1}$ vanishes whenever adjacent steps share the same $a$. A change of $\lambda$ can therefore alter the current-token term while leaving the past-token weights nearly unchanged.

\section{The continuum limit and finite memory}\label{sec:refinement}

This section asks when $\lambda$ can matter. On a dense token grid it cannot, since every $\lambda$ gives the same limit (Theorem~\ref{thm:volterra}). At a short memory horizon it can. A constant, nonrotating head with $\lambda=1/2$ never puts most of its squared weight on the previous token, while a short-horizon head with small $\lambda$ can (Proposition~\ref{prop:lagone}). Appendix~\ref{app:proofs} gives the proofs and a numerical check.

We begin with a single layer at a fixed input. There, refining the grid means splitting each step $k$ into $n$ substeps of size $\Delta_k/n$. The substeps keep the state-inputs and $C$ projections at the original tokens, and their state transitions multiply to the original $\alpha_k\mathbf R_k$. If two rules share the substep state-inputs and state transitions and keep step $k$'s difference $\lambda_k-\mu_k$ on each of its substeps, their contributions at interior substeps telescope in \eqref{eq:difference}. Both $\mathcal D_i$ and $\mathcal O_i$ at the original tokens then shrink to exactly $1/n$ of their unrefined values, so the effect of $\lambda$ vanishes as $n$ grows. We use this identity to test the whole network and to check the released Triton kernel.

\subsection{A common Volterra limit}

Refining the token grid across depth is a different operation: it adds token states that evolve through the head's own output. We study a single-head SISO surrogate over depth $t\in[0,t_f]$ and sequence position $s\in[0,1]$. It builds on continuous-depth models of residual networks \citep{chen2018neuralode}, particle models of attention \citep{geshkovski2023emergence} and token dynamics of selective SSMs across depth \citep{vo2024demystifying,nguyen2025tokenspheres}. The RMSNorm pre-norm motivates token states $u_i(t)$ on the RMS sphere $\Srms=\{u\in\R^d:\|u\|^2=d\}$ evolving according to
\begin{equation}\label{eq:surrogate}
\frac{du_i}{dt}=P_{u_i}S_o\,\bm y_i[u(t)],\qquad P_u=I-\frac{uu^\top}{d},
\end{equation}
where $P_u$ projects onto the tangent space of $\Srms$ at $u$ and $S_o\in\R^{d\times P}$ is the output projection to the model dimension. In $\bm y_i[u]$, the SSM input $\bm x_i=W(s_i,u_i)$ and the $B$ and $C$ projections $B_i=B(s_i,u_i)$ and $C_i=C(s_i,u_i)$ of Section~\ref{sec:setup} are evaluated on the states $u_i(t)$ at the token positions $s_i$ defined below. The step sizes, decays, rotation angles and $\lambda$ are computed from the initial states and frozen over depth. The surrogate omits the skip term, gate and SwiGLU block.

To refine the grid, we replace $\Delta_i$ in \eqref{eq:quadrature} by the normalized steps $\delta_i=\Delta_i/\sum_{k\le T}\Delta_k$, placing token $i$ at $s_i=\sum_{k\le i}\delta_k\in[0,1]$ with mesh width $\delta_{\max}=\max_i\delta_i$ and cumulative profiles $H_T(s_i)=H_i$ and $\bm\Phi_T(s_i)=\bm\Phi_i$.

\begin{theorem}\label{thm:volterra}
Let $T\to\infty$ along meshes with $\delta_{\max}\to0$, and assume the following.

\textup{(A1)}~\emph{\underline{Gradual decay and rotation}:} $H_T$ and $\bm\Phi_T$ converge at the mesh points to a nondecreasing Lipschitz profile $H$ and a Lipschitz profile $\bm\Phi$, with $H(0)=0$, $\bm\Phi(0)=0$ and combined mesh-point error
\[
\varepsilon_T=\max_i|H_T(s_i)-H(s_i)|+\max_i\|\bm\Phi_T(s_i)-\bm\Phi(s_i)\|\longrightarrow0.
\]
\textup{(A2)}~\emph{\underline{Smooth inputs}:} $B$, $C$ and $W$ are bounded and Lipschitz in $(s,u)$, $u_i(0)\in\Srms$ for every $i$, and the initial error $e_T(0)=\max_i\|u_i(0)-u_0(s_i)\|\to0$ for a Lipschitz field $u_0:[0,1]\to\Srms$. 

Then, uniformly over $\lambda\in[0,1]^T$ held fixed over depth, the solutions of \eqref{eq:surrogate} converge at the mesh points to the unique solution $u:[0,1]\times[0,t_f]\to\Srms$ of
\begin{equation}\label{eq:volterra}
\partial_tu(s,t)=P_{u(s,t)}S_o\int_0^s\kappa\bigl(s,\sigma,u(s,t),u(\sigma,t)\bigr)\,W\bigl(\sigma,u(\sigma,t)\bigr)\,d\sigma,
\end{equation}
with $u(\cdot,0)=u_0$ and kernel
\[
\kappa(s,\sigma,u,w)=C(s,u)^\top U(s,\sigma)B(\sigma,w),\qquad U(s,\sigma)=e^{-[H(s)-H(\sigma)]}\mathbf R\bigl(\bm\Phi(s)-\bm\Phi(\sigma)\bigr).
\]
More precisely, for constants $L$ and $K$ independent of $T$ and $\lambda$,
\[
\max_{t\le t_f}\max_i\|u_i(t)-u(s_i,t)\|\le e^{Lt_f}\bigl\{e_T(0)+Kt_f(\delta_{\max}+\varepsilon_T)\bigr\}.
\]
\end{theorem}

The limit \eqref{eq:volterra} does not involve $\lambda$. Its kernel is set by decay, rotation and the $B$ and $C$ projections. On the token grid, (A1) means that each step decays and rotates the state only slightly, since $q_i\le L_H\delta_i+2\varepsilon_T\to0$ for the Lipschitz constant $L_H$ of $H$, and likewise for $\|\bm\omega_i\|$. The local memory horizon $1/q_i$ then grows, and mesh refinement keeps each diagonal entry $\lambda_i\delta_i$ at most $\delta_{\max}$. (A2) keeps adjacent state-inputs close. The two assumptions together bound the difference from the exponential-Euler sum ($\lambda\equiv1$) in \eqref{eq:difference} by $O(\delta_{\max}+\varepsilon_T)$ uniformly in $\lambda$. A Gr\"onwall estimate then propagates agreement across depth, and numerical errors shrink at a rate near $T^{-1}$. By Theorem~\ref{thm:volterra}, $\lambda$ can therefore have a large effect only if (A1) or (A2) fails, that is, only with short memory horizons, large rotations per step or sharp jumps between adjacent tokens. 

\subsection{At a finite memory horizon}

At a short memory horizon, $\lambda$ decides how much of the SSM output comes from the current token. The current-token weight in turn limits how strongly a head can pass the previous token forward. Such horizons occur on the token grid: every released model contains early heads whose memory horizon is below two tokens.

Consider a head with constant $\Delta$, $q>0$ and $\lambda$, no rotation, and fixed $B$ and $C$ projections with $B^\top C\ne0$. With $\alpha=e^{-q}$, \eqref{eq:quadrature} gives
\[
\bm y_i=(B^\top C)\sum_{m=0}^{i-1}k_m\bm x_{i-m},\qquad
k_0=\lambda\Delta,\quad k_m=\Delta\alpha^m\quad(m\ge1),
\]
where $k_m$ weights the input $m$ tokens back, at lag $m$. Varying $\lambda$ changes only the lag-zero weight $k_0$. At every past lag $m\ge1$, the weights from the two adjacent steps sum to $[\lambda+(1-\lambda)]\Delta\alpha^m=\Delta\alpha^m$, independent of $\lambda$. Under a constant input, the SSM output approaches $(B^\top C)G_\lambda\bm x$ with gain $G_\lambda=\sum_{m\ge0}k_m=\Delta\bigl(\lambda+\alpha/(1-\alpha)\bigr)$. Since $G_\lambda-G_\mu=\Delta(\lambda-\mu)$, normalizing by the continuous-time gain $\Delta/q$ at decay rate $q/\Delta$ gives $(G_\lambda-G_\mu)/(\Delta/q)=(\lambda-\mu)q$. Raising $\lambda$ from $0$ to $1/2$ shifts this normalized gain by $0.5$ at a one-token horizon ($q=1$) but only about $0.008$ at a 64-token horizon ($q=1/64$).
A head that passes the previous token forward needs most of its weight at lag one. Let $p_1=k_1^2/\sum_{m\ge0}k_m^2$ denote this lag-one share of the squared weights.

\begin{proposition}\label{prop:lagone}
For a nonrotating head with constant $\Delta$, $q>0$, $\lambda\in[0,1]$, and fixed $B$ and $C$ projections ($B^\top C\ne0$), the lag-one share satisfies $p_1\le1/(1+\lambda)^2$. For $\lambda>0$, equality holds if and only if $\alpha^2=\lambda/(1+\lambda)$. For $\lambda=0$, $p_1=1-\alpha^2<1$, which tends to one as $q\to\infty$.
\end{proposition}

At $\lambda=1/2$, such a head therefore puts at most $4/9$ of its squared weight on the previous token. Lowering $\lambda$ shrinks $k_0=\lambda\Delta$ while leaving $k_m$ unchanged for $m\ge1$, so $p_1$ can approach one at $\lambda=0$ as $q$ grows. A head with $\lambda$ fixed at $1/2$ can exceed the $4/9$ ceiling only through rotation or through $B$, $C$ or decays that vary across tokens. One way is to make $B_i$ nearly orthogonal to $C_i$ while the lag-one coefficient, which pairs $C_i$ with $B_{i-1}$, stays large.

\section{The rule on the token grid}\label{sec:native}

The continuum limit says where the rule can matter. The released checkpoints show where it does.

\subsection{Where the reset cost sits}

We reset $\lambda$ to $1/2$ one layer at a time and measure the loss increase on PG-19 books \citep{rae2020compressive}. The cost concentrates in early layers (Figure~\ref{fig:headline}a--d), so we select one early layer per model on PG-19: layers 2, 3, 1 and 3 for 187M, 443M, 893M and 1.5B. Each reset intervention below applies the current-token change $\mathcal D_i$ or the past-token change $\mathcal O_i$ of \eqref{eq:presentpast}, or resets $\lambda$ to $1/2$ and restores the learned current-token term in a set of heads. All $\lambda$ interventions keep the learned skip term, gate and output projection. The text uses 443M as the running example. Figure~\ref{fig:headline} and Table~\ref{tab:dial} give all four models. Appendix~\ref{app:exp} gives the evaluation setup, the full tables and further checks.

 Applying only the current-token change in the selected layer reproduces the LAMBADA drop of the reset in every model. In 443M, the reset costs 18.8 points and the current-token change alone costs 18.9 (Figure~\ref{fig:headline}). Restoring the learned current-token term recovers over 90\% of the summed one-layer PG-19 loss increase (Figure~\ref{fig:headline}g,h). The one-layer cost therefore sits in the current-token term. The past-token change barely affects LAMBADA in one layer (0.3 points in 443M). In all layers at once, it costs 13--15 points, so its effect builds up across depth (Figure~\ref{fig:headline}f).

To find the heads that carry the one-layer cost, we replace each $\lambda_i$ with its head's mean $\bar\lambda$, computed on separate validation books. This replacement costs far less than $\lambda=1/2$ in every model, 12\% as much in 443M. Most of the cost therefore comes from shifting the mean $\lambda$ of each head to $1/2$, not from removing its variation across tokens. We call heads with $\bar\lambda<0.2$ \textbf{low-$\bm\lambda$} heads and test whether they carry the cost. In layer 3 of 443M, 9 of the 32 heads are low-$\lambda$, and restoring their current-token term lowers the loss increase of the reset from 0.52 to 0.007 nats per token. In every model, low-$\lambda$ heads give most of the recovery from restoring all heads. The same holds without reselection on WikiText-103 \citep{merity2017pointer}, Python code and FineWeb-Edu \citep{penedo2024fineweb}. Setting the low-$\lambda$ heads of the selected layer to $\lambda=0$, which lies within $0.2$ of their mean, costs over 20 times less than $\lambda=1/2$, 560 times less in 443M. The cost comes from raising $\lambda$ to $1/2$.

\begin{figure}[!htbp]
\centering
\includegraphics[width=\textwidth]{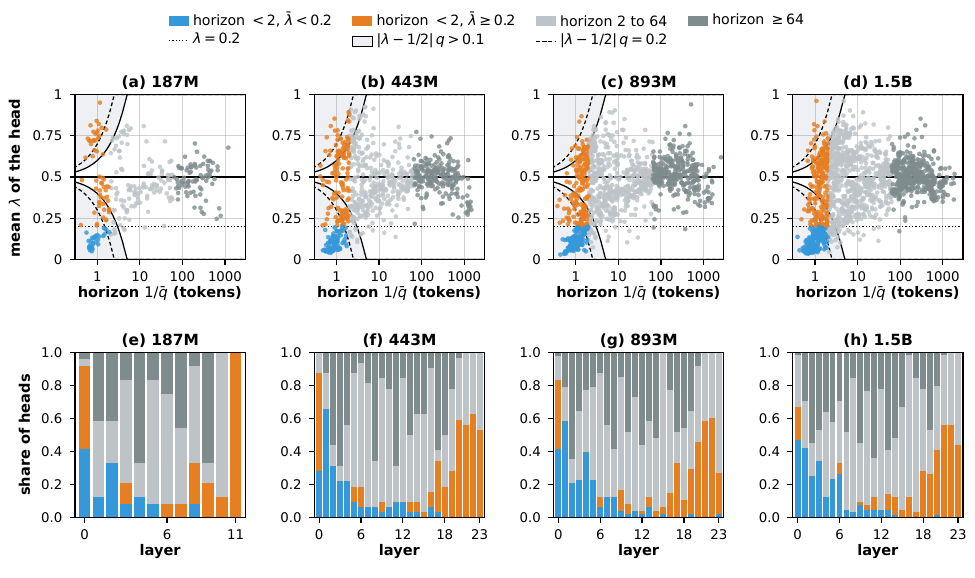}
\caption{Early short-horizon heads tend to have small mean $\lambda$, while most long-horizon heads stay near $1/2$. (a--d) Mean $\lambda$ and memory horizon $1/\bar q$ of each head on the first 16 books of PG-19 set 2. The shaded region marks normalized gain changes above 0.1 under $\lambda=1/2$ for a constant, nonrotating head, and the dashed curve marks a change of 0.2. (e--h) Share of heads of each group in each layer.}
\label{fig:dial}
\vspace{-0.2in}
\end{figure}

Most low-$\lambda$ heads also have a memory horizon below two tokens. We summarize a head's memory horizon by $1/\bar q$, where $\bar q$ is its per-step decay $q_i$ averaged over all tokens and books. We call heads \textbf{short-horizon} when $1/\bar q<2$ and \textbf{long-horizon} when $1/\bar q\ge64$. Low-$\lambda$ heads make up about 12\% of all heads, and 78--85\% of them are short-horizon in every model, 77 of 91 in 443M. Short-horizon heads have small $\bar\lambda$ mainly in early layers (0--3 in 187M, 0--6 in the others), and most long-horizon heads keep $\bar\lambda$ within $0.1$ of $1/2$ (Figure~\ref{fig:dial}).

\begin{table}[!htbp]
\centering
\caption{Head-group interventions on PG-19, as medians over 32 books (16 for $R$). $\Delta L$ is the selected layer's loss increase in nats per token under a full reset and after restoring the learned current-token term in low-$\lambda$ heads. Control costs are relative to the cost of the current-token change, and Appendix~\ref{app:exp} defines all ratios.}
\label{tab:dial}
\small
\input{tables/tab_dial.tex}
\end{table}

 In these short-horizon low-$\lambda$ heads, the current-token term partially cancels the skip term $D\bm x_i$. In 443M, the squared norm of their gated sum is $R=0.18$ times the sum of their separate squared norms (Table~\ref{tab:dial}). Raising $\lambda$ to $1/2$ upsets this balance. Two controls test whether the size of the change explains its cost. They replace the current-token change in the low-$\lambda$ heads of layers 1--4 (1--2 in 187M) by other changes of the same size. A random change with the same norm at every token costs far less in every model, 0.25 times as much in 443M. A change of the skip parameter $D$ with the same pooled norm and sign costs more, 2.3 times as much in 443M. The skip term also acts only on the current token, so both costly changes alter how strongly the current token enters the output of a head.

\subsection{Where the assumptions of Theorem~\ref{thm:volterra} fail}

On a finite grid, (A1) and (A2) of Theorem~\ref{thm:volterra} ask for per-step decay, rotation and input jumps of order $\delta_{\max}\approx10^{-3}$. We check them on 16 PG-19 windows.

(A1) fails in early short-horizon heads. A memory horizon below two tokens means a mean per-step decay above $1/2$, hundreds of times $\delta_{\max}$. These heads reach a total rotation $\sum_i\|\bm\omega_i\|$ above $5{,}000$ and a cumulative decay $H_T(1)$ of $1{,}967$ in 443M, against $8.1$ in long-horizon heads. Longer contexts do not repair (A1), since Mamba-3's decay floor makes the total decay grow with the sequence length. (A2) fails in every head group. The layer inputs jump by order one between adjacent tokens, and the relative state-input jumps are about $2{,}700$ times the normalized step $\delta_i$.

To test whether these failures let $\lambda$ matter, we compare the SSM outputs of the learned rule and $\lambda=1/2$ in the selected layer's low-$\lambda$ heads, with and without each assumption enforced. On unmodified inputs, their relative difference is of order one (3.59 in 443M). Smoothing the inputs alone leaves it of order one. Rescaling decay and rotation alone, so that $H_T(1)\le1$, lowers it by an order of magnitude. Doing both lowers it more than a hundredfold (0.022 in 443M). The checkpoints violate both assumptions, so $\lambda$ still changes their SSM output.

In summary, the reset cost sits in the current-token term of low-$\lambda$ heads in early layers. Low-$\lambda$ heads make up about 12\% of all heads and are mostly short-horizon. Early short-horizon heads break (A1), so they sit where Theorem~\ref{thm:volterra} allows $\lambda$ to matter. Proposition~\ref{prop:lagone} shows that a head with constant parameters needs a small current-token weight to put most of its squared weight on the previous token. Yet models trained with $\lambda$ fixed at $1/2$ reach nearly the same perplexity as models that learn it. How do early short-horizon heads keep the current token out when training fixes its weight at $1/2$?

\section{The rule in training}\label{sec:training}

The four training rules reach nearly the same validation loss, but their heads keep the current token out of the SSM output in different ways. In three matched 187M seeds, fixing $\lambda$ at $1/2$ in both the state update and the current-token term raises validation loss by only $0.003$--$0.011$ nats per token relative to learning it. Fixing it in the state update or in the current-token term alone changes the loss by at most $0.008$ nats (Table~\ref{tab:training_summary}). Appendix~\ref{app:training} gives the training setup, the tests and the results for each seed.

\subsection{What the trained heads learn}\label{sec:learn}

\begin{figure}[!htbp]
\centering
\includegraphics[width=\linewidth]{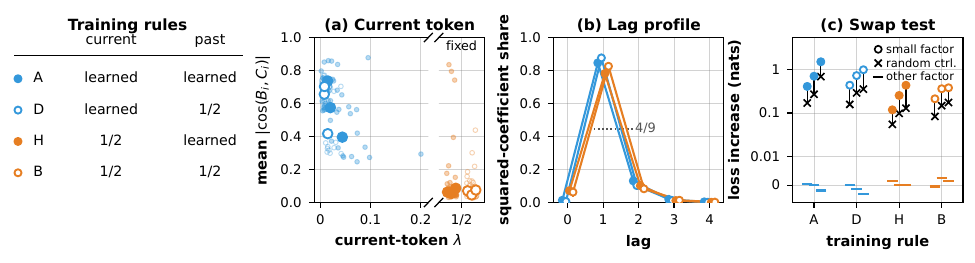}
\caption{Under all four training rules the final short-horizon heads of layers 1--3 emphasize the previous token. (a) Current-token weight and mean $|\cos(B_i,C_i)|$ of each head (dots) and run medians (large markers), with rules H and B jittered at their fixed $1/2$. (b) Shares of the squared coefficients of the SSM output by lag, normalized over lags 0--16 and shown for lags 0--4 (medians and ranges over three seeds), with the $4/9$ ceiling of Proposition~\ref{prop:lagone} (dotted). (c) PG-19 loss increase in the swap test (linear below 0.01 nats). Circles: the small factor takes its value in the paired rule. Crosses: a norm-matched random vector. Dashes: the other factor is swapped.}
\label{fig:training}
\vspace{-0.1in}
\end{figure}

To separate the two roles of $\lambda$, let $\ell_i$ govern the recurrent state update and let the current-token weight $c_i$ scale the current-token term of the SSM output:
\[
\bm h_i^\ell=\alpha_i\mathbf R_i\bm h_{i-1}^\ell+\alpha_i\Delta_i(1-\ell_i)\mathbf R_i\bm v_{i-1}+\Delta_i\ell_i\bm v_i,\qquad
\bm y_i^{\ell,c}=(\bm h_i^\ell)^\top C_i+\Delta_i(c_i-\ell_i)F_{ii}.
\]
The correction $\Delta_i(c_i-\ell_i)F_{ii}$ changes the current-token term of the SSM output while leaving $\bm h_i^\ell$ unchanged. Our $2\times2$ design compares four rules (Figure~\ref{fig:training}). Rule A learns both weights ($c_i=\ell_i=\lambda_i$), rule D learns the current-token weight only ($c_i=\lambda_i$, $\ell_i=1/2$), rule H learns the past-token weights only ($c_i=1/2$, $\ell_i=\lambda_i$), and rule B fixes both at $1/2$.

We train each 187M model on 1.46B FineWeb-Edu tokens \citep{penedo2024fineweb}, sharing initialization and data order within each seed, and select the heads in layers 1--3 that finish training with a short horizon. Head statistics and interventions use held-out PG-19 books. The lag-one and swap tests below change the selected heads of one layer at a time, and we sum the loss increases over layers 1--3.

Under all four rules, the selected short-horizon heads emphasize the previous token (Figure~\ref{fig:training}b). In every rule and seed, the median lag-one share over lags 0--16 exceeds $0.75$ and the lag-zero share stays below $0.1$. Including the skip parameter $D$ in the lag-zero coefficient leaves the lag-zero share small under all four rules. Removing the lag-one contribution raises PG-19 loss by more than twice the cost of a norm-matched random perturbation in every run (Table~\ref{tab:training_summary}). Under rules B and H, the lag-one share exceeds the $4/9$ ceiling of Proposition~\ref{prop:lagone}, even though the current-token weight is fixed at $1/2$. Heads under rules B and H keep the current token out by making $B_i$ and $C_i$ nearly orthogonal in every seed. Their lag-one coefficient pairs $C_i$ with the rotated $B_{i-1}$ and stays large. Heads under rules A and D instead drive the learned current-token weight $\lambda$ near zero, and the $B_i$ and $C_i$ of the same token end up positively aligned (Figure~\ref{fig:training}a).

 We test which factor keeps the lag-zero coefficient $c_i\Delta_iB_i^\top C_i$ small in each rule. We call the factor that a rule keeps near zero its small factor: $\lambda$ in rules A and D, and $B_i^\top C_i$ in rules H and B. The test pairs rules that differ only in whether the current-token weight is learned, A with H and D with B. It changes one factor at a time and holds the recurrent state and the past-token weights fixed (Figure~\ref{fig:training}c, Table~\ref{tab:training_summary}). The swap sets $c_i=1/2$ in rules A and D. In rules H and B, it replaces $B_i^\top C_i$ by $\|B_i\|\|C_i\|$ times the paired run's median signed cosine between $B_i$ and $C_i$. This swap raises PG-19 loss by more than twice the cost of a random change of the same norm in every run. Swapping the other factor changes loss by less than $0.004$ nats in every run, because the small factor already holds their product near zero.

\begin{table}[!htbp]
\centering\small
\caption{The four training rules at 187M, as medians over three seeds. Loss gap is the validation loss minus that of rule A. The swap and lag-one tests give summed PG-19 loss increases in nats per token.}
\label{tab:training_summary}
\input{tables/tab_training_summary.tex}
\end{table}

\subsection{During training}

\begin{figure}[!htbp]
\centering
\includegraphics[width=0.98\linewidth]{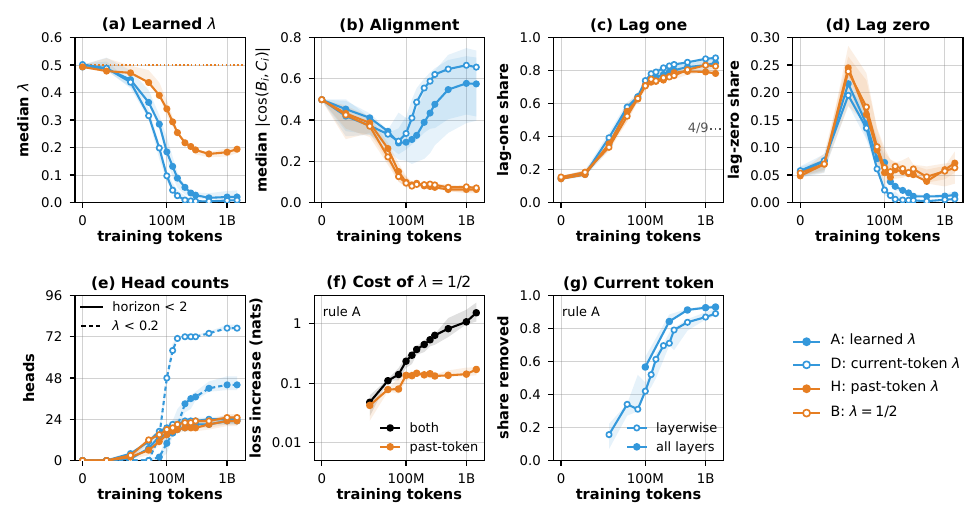}
\caption{Every rule first lowers the alignment of $B_i$ and $C_i$. Rules A and D then drive $\lambda$ below $0.1$ by 200M tokens and realign $B_i$ and $C_i$, while rules B and H keep them nearly orthogonal ($|\cos(B_i,C_i)|\approx0.1$ from 100M tokens on). (a--d) Final short-horizon heads of layers 1--3 tracked from initialization. In (a), H shows its past-token $\lambda$. (e) Number of the 96 heads in layers 0--3 that are short-horizon (solid, all four rules) or low-$\lambda$ (dashed, A and D) at two consecutive checkpoints. (f) Summed one-layer PG-19 loss increase of rule A under $\lambda=1/2$, for the full change and the past-token change. (g) Share of the cost removed by restoring the learned current-token term, layerwise and across all layers together. Lines and bands are medians and ranges over three seeds on a logarithmic token axis with initialization at 0. The dotted lines mark $\lambda=1/2$ in (a) and the $4/9$ ceiling of Proposition~\ref{prop:lagone} in (c).}
\label{fig:evolution}
\end{figure}

Tracking the heads of Section~\ref{sec:learn} back to initialization shows how the two mechanisms form (Figure~\ref{fig:evolution}a--d). Early in training, every rule lowers the initial alignment of $B_i$ and $C_i$ ($|\cos(B_i,C_i)|$ near $0.5$). The rules then separate. In rules A and D, $\lambda$ drops below $0.1$ by 200M tokens, and the alignment of $B_i$ and $C_i$ rises again as $\lambda$ declines. The short horizon usually comes first: across all 96 heads of layers 0--3, 60--83\% of the heads that become both short-horizon and low-$\lambda$ reach the short horizon before the small $\lambda$. In rules B and H, $B_i$ and $C_i$ stay nearly orthogonal. On both paths, the heads put over 70\% of their squared weight on the previous token by 125M tokens.

At 25M tokens, the cost of resetting $\lambda$ to $1/2$ in rule A is small and comes mostly from the past-token change. As $\lambda$ falls, the cost grows and more of it comes from the current-token change. At the end of training, restoring the learned current-token term removes about 90\% of the summed one-layer cost. It removes over 90\% of the cost of resetting all layers together, as in the released checkpoints (Figure~\ref{fig:evolution}f,g).

In summary, a head with its current-token weight fixed at $1/2$ can keep the current token out by making $B_i$ and $C_i$ nearly orthogonal. Fixing $\lambda$ before training therefore costs little. Heads that learn the current-token weight realign $B_i$ and $C_i$ and depend on the small $\lambda$ alone, so a reset after training costs far more than training with $\lambda$ fixed.

\section{Related work}\label{sec:related}

Separating the exact state transition from input quadrature is classical for exponential integrators \citep{hochbruck2010exponential}. Structured SSMs discretize continuous-time systems \citep{gu2020hippo,gu2021lssl,gu2022s4}, selective SSMs make step sizes input-dependent \citep{gu2023mamba,dao2024mamba2}, and Mamba-3 introduces the learned exponential-trapezoidal rule and its structured mask \citep{lahoti2026mamba3}. Our decomposition \eqref{eq:presentpast} turns the split between state transitions and mask weights into an exact intervention on trained models.

Along the sequence, selective SSMs have been analyzed as controlled differential equations driven by the input path \citep{kidger2020neuralcde,cirone2024theoretical}. Across depth, SSMs and transformers have been studied as continuous-depth token flows \citep{chen2018neuralode,vo2024demystifying,nguyen2025tokenspheres} and mean-field limits of attention \citep{geshkovski2023emergence,karagodin2024clustering,rigollet2025meanfield,castin2025unified}. Classical Volterra equations fix the kernel, and neural integro-differential equations parameterize it directly \citep{gripenberg1990volterra,zappala2023neuralide}. In Theorem~\ref{thm:volterra}, the SSM itself sets a causal kernel that does not depend on $\lambda$.

Knockout and attribution studies trace token-to-token flow in Mamba \citep{endy2025mambaknockout,pitorro2025latim,hossain2025selectivememory}. Context-extension methods drop tokens \citep{benkish2024decimamba} or rescale decays and step sizes \citep{lu2025mambamodulation,shen2026upi}. LaTIM folds the skip term into the current-token term \citep{pitorro2025latim}, while Section~\ref{sec:setup} separates the current-token term $\gamma_iF_{ii}$ from the skip term $D\bm x_i$ and past tokens.

The cost of ablating a component of a trained network depends on what the rest of the network does in its place. In GPT-2 small, backup name mover heads take over when the name mover heads are ablated \citep{wang2023ioi}, and in Chinchilla another attention layer compensates for an ablated one \citep{mcgrath2023hydra}. This self-repair also occurs after single-head ablations across the pretraining distribution, partly through changes in the final LayerNorm scale \citep{rushing2024selfrepair}. Such compensation makes an ablation understate a component's role. With further training, the rest of the network can replace a component. Removing LayerNorm from a trained GPT-2 raises its loss, and fine-tuning after each removal brings the loss back close to the original in every model size up to GPT-2 XL \citep{heimersheim2024layernorm,baroni2026layernorm}. Section~\ref{sec:native} resets $\lambda$ in the released checkpoints without retraining, and Section~\ref{sec:training} compares learned $\lambda$ with $\lambda$ fixed before training.

Passing the previous token forward recurs across sequence architectures. H3 uses an explicit shift SSM \citep{fu2023h3}, and previous-token heads underpin transformer induction circuits \citep{elhage2021framework}. In Mamba-1, \citet{wang2025universality} find an off-by-one pattern where the short convolution shifts token information into the next state. In small Mamba-1 models on synthetic tasks, \citet{arora2025mechanistic} show that removing this convolution breaks associative recall. They find that the models still learn associative treecall through a two-layer induction mechanism, with lower accuracy. Mamba-3 removes the short convolution, and its authors credit the $B$ and $C$ biases and the trapezoidal rule with replacing it \citep[Section~3.4]{lahoti2026mamba3}. In our training runs, short-horizon heads pass the previous token forward under both learned and fixed current-token weights, using either a small $\lambda$ or nearly orthogonal $B_i$ and $C_i$.

\section{Discussion}\label{sec:discussion}

A parameter can matter little when it is fixed before training, because other parameters compensate. The same parameter can still carry the predictions of a model that learned it. In heads trained with $\lambda$ fixed at $1/2$, nearly orthogonal $B_i$ and $C_i$ do the work of a small $\lambda$. Training with a fixed $\lambda$ and resetting a learned $\lambda$ therefore answer different questions.

Mamba-3 motivates $\lambda$ through numerical integration. Yet in the released checkpoints, $\lambda$ acts in short-horizon heads mainly as a weight on the current token. These heads break both assumptions of Theorem~\ref{thm:volterra}, under which the Volterra limit removes every effect of $\lambda$.

These findings point to two design observations. First, rule B has a higher validation loss than rule H in every seed and than rule D in two of the three seeds, by up to $0.007$ nats (Appendix~\ref{app:training}). Rule D has 1.6 to 1.9 times as many heads that become low-$\lambda$ in layers 0--3 as rule A. In the released checkpoints, replacing each $\lambda_i$ with its head's mean $\bar\lambda$ costs 10--19\% as much as $\lambda=1/2$ (Table~\ref{tab:dial}). The next step is therefore to test at scale a current-token weight that does not depend on the input, like the bonus term of RWKV \citep{peng2023rwkv}.

Second, for training-free context extension \citep{lu2025mambamodulation,shen2026upi}, rescaling $A_i$ at a fixed layer input modifies the state transitions and leaves the mask weights $\gamma_i$ and $\eta_j$ intact. Rescaling $\Delta_i$ instead alters both the state transitions and the balance between the current-token term and the skip term $D\bm x_i$.

Our theory treats a single-head SISO surrogate with step sizes, state transitions and $\lambda$ frozen over depth. The training runs span 1.46B tokens at 187M parameters. The theory remains to be extended to state-dependent transitions over depth, the training analysis to longer runs, and both to rank-$r$ multi-input, multi-output (MIMO) Mamba-3 layers. In these layers, keeping the current token out at fixed $\lambda=1/2$ through the projections means suppressing $B_i^\top C_i\in\R^{r\times r}$, which constrains $r^2$ inner products per token, not one.

\section*{Acknowledgments}
F.Y.\ is grateful for partial support from seed funding by the Center for Emerging Artificial Intelligence Systems at the University at Albany.

\bibliographystyle{tmlr}
\bibliography{refs,refs_native}

\appendix
\numberwithin{lemma}{section}
\numberwithin{equation}{section}
\numberwithin{table}{section}
\numberwithin{figure}{section}

\section{The Mamba-3 layer}\label{app:setup}

In the four released SISO checkpoints \citep{lahoti2026mamba3}, each of the $L$ network blocks ($L=12$ in 187M and $L=24$ in the three larger models) pairs a Mamba-3 layer with a SwiGLU block. Both have an RMSNorm pre-norm and residual connection. Let $u_i=\operatorname{RMSNorm}_1(r_i)$ be the pre-normed input to the Mamba-3 layer from residual state $r_i$. A single linear projection produces all layer inputs:
\[
(\bm z_i,\bm x_i,\widetilde B_i,\widetilde C_i,
 \widehat{\bm\Delta}_i,\widehat{\bm A}_i,\widehat{\bm\lambda}_i,\bm\vartheta_i)
 =W_{\rm in}u_i.
\]
Here $\bm x_i$ and $\bm z_i$ split across $H$ heads with $\bm x_{i,h},\bm z_{i,h}\in\R^P$, the shared projections $\widetilde B_i,\widetilde C_i\in\R^N$ pass through separate RMSNorm layers (BCNorm) before head-specific biases $b^B_h,b^C_h\in\R^N$ are added, and $\widehat{\bm\Delta}_i,\widehat{\bm A}_i,\widehat{\bm\lambda}_i\in\R^H$:
\[
\begin{aligned}
 B_{i,h}&=\operatorname{RMSNorm}_B(\widetilde B_i)+b^B_h, &
 C_{i,h}&=\operatorname{RMSNorm}_C(\widetilde C_i)+b^C_h,\\
 \Delta_{i,h}&=\operatorname{softplus}(\widehat\Delta_{i,h}+b^\Delta_h), &
 \lambda_{i,h}&=\operatorname{sigmoid}(\widehat\lambda_{i,h}),\\
 A_{i,h}&=-\max\{\phi(\widehat A_{i,h}),A_{\rm floor}\}, &&
\end{aligned}
\]
with $A_{\rm floor}=10^{-4}$, $\phi(a)=1+a$ for $a\ge0$, and $\phi(a)=1/(1-a)$ for $a<0$. The released Triton kernel implements rotation by rotating $B_{i,h}$ and $C_{i,h}$ on 32 of the 64 coordinate pairs of the $N=128$ state through cumulative angles $\bm\theta_{i,h}=\sum_{k\le i}\pi\Delta_{k,h}\tanh(\bm\vartheta_k)$. Since $(\mathbf R(\bm\theta_{i,h})C_{i,h})^\top\mathbf R(\bm\theta_{j,h})B_{j,h}=C_{i,h}^\top\mathbf R(\bm\theta_{j,h}-\bm\theta_{i,h})B_{j,h}$, this matches \eqref{eq:recurrence} with $\bm\omega_{i,h}=-\pi\Delta_{i,h}\tanh(\bm\vartheta_i)$ on the rotated pairs and zero on the rest.

Each head adds the skip term $D_h\bm x_{i,h}$ to $\bm y_{i,h}$, gates with $\operatorname{SiLU}(\bm z_{i,h})$, and projects back to the model dimension (with no intermediate normalization in the released checkpoints):
\[
\bm o_i=W_{\rm out}\operatorname{concat}_{h=1}^H
\bigl[(\bm y_{i,h}+D_h\bm x_{i,h})\odot\operatorname{SiLU}(\bm z_{i,h})\bigr].
\]
Throughout our experiments, a layer's output refers to $\bm o_i$ prior to the residual addition and SwiGLU block. At a fixed layer input, changing $\lambda$ leaves $D_h$, $\bm z_{i,h}$ and $W_{\rm out}$ unchanged, so \eqref{eq:presentpast} and \eqref{eq:identity} carry over directly to the layer output $\bm o_i$.

The released kernel rounds $C$, $B$, $\bm x$, $\bm z$ and the $\lambda$ and angle inputs to BF16, uses fast $\tanh$, $\sin$ and $\cos$ approximations, and accumulates the recurrence in FP32. Two independent float64 implementations of \eqref{eq:recurrence} agree to $10^{-15}$ and differ from the fused kernel by 0.3\%--0.7\% in relative root mean square. At $n=1$ our wrapper reproduces the released logits bit for bit, and $n=8$ subdivision divides the $\lambda=1/2$ change by 8 in float64 and by 7.993 (443M) and 7.994 (1.5B) in the fused kernel.

\section{Proofs}\label{app:proofs}

Vector norms are Euclidean and matrix norms are operator norms. Since rotations are orthogonal and $0<\alpha_k<1$, $\|U_{ij}\|\le1$. We set $F_{i0}=0$. Subdivision and constant-head results use unnormalized steps $\Delta_k$, whereas the finite-mesh bound and Theorem~\ref{thm:volterra} use normalized steps $\delta_k=\Delta_k/S$ with $S=\sum_k\Delta_k$ (multiplying outputs and bounds by $S$ recovers unnormalized steps).

\subsection{Proof of \texorpdfstring{Theorem~\ref{thm:volterra}}{Theorem 1}}\label{app:volterra-proof}\label{app:subdivide}\label{app:finite}

We first record the subdivision identity and finite-mesh bound at a fixed layer input. Subdivide each step $k$ into $n$ substeps with step size $\Delta_k/n$, decay $q_k/n$, angles $\bm\omega_k/n$, and step $k$'s rule parameter, keeping the state-inputs $\bm v_{k-1},\bm v_k$ at the endpoints and arbitrary shared interior samples. Writing $F_{i,k,r}$ for sample $r$ of step $k$, the substep transitions compose to $\alpha_k\mathbf R_k$, so $F_{i,k,0}=F_{i,k-1}$ and $F_{i,k,n}=F_{ik}$. By \eqref{eq:difference}, step $k$'s substeps telescope to $\frac{\Delta_k}{n}(\lambda_k-\mu_k)\sum_{r=1}^n(F_{i,k,r}-F_{i,k,r-1})=\frac{\Delta_k}{n}(\lambda_k-\mu_k)(F_{ik}-F_{i,k-1})$, giving
\begin{equation}\label{eq:identity}
\bm y_i^{\lambda,(n)}-\bm y_i^{\mu,(n)}=\frac1n\bigl(\bm y_i^\lambda-\bm y_i^\mu\bigr).
\end{equation}
Because $a_{k,r}=\Delta_k(\lambda_k-\mu_k)/n$ has zero interior jumps, both $\mathcal D_i$ and each term of $\mathcal O_i$ scale by $1/n$, whereas rescaling substep decays by $c\ne1$ leaves $U_{ij}=e^{-c(H_i-H_j)}\mathbf R(\bm\Phi_i-\bm\Phi_j)$ altered for every $n$.

Under normalized steps $\delta_k$, \eqref{eq:difference} with $|\lambda_k-\mu_k|\le1$ and $F_{i0}=0$ gives $\max_i\|\bm y_i^\lambda-\bm y_i^\mu\|\le\delta_{\max}(M_F+V_F)$ for all $\lambda,\mu\in[0,1]^T$, where $M_F=\max_i\|F_{i1}\|$ and $V_F=\max_i\sum_{k=2}^i\|F_{ik}-F_{i,k-1}\|$. With $M_C=\max_i\|C_i\|$, $M_v=\max_k\|\bm v_k\|$, $V_v=\sum_{k\ge2}\|\bm v_k-\bm v_{k-1}\|$ and $V_A=\sum_{k\ge2}\|I-\alpha_k\mathbf R_k\|$, the relation $U_{i,k-1}=U_{ik}\alpha_k\mathbf R_k$ gives $F_{ik}-F_{i,k-1}=\bigl[U_{ik}\bigl((\bm v_k-\bm v_{k-1})+(I-\alpha_k\mathbf R_k)\bm v_{k-1}\bigr)\bigr]^\top C_i$. Since $\|U_{ij}\|\le1$, we obtain $M_F\le M_CM_v$ and $V_F\le M_C(V_v+M_vV_A)$. Under $n$-fold subdivision with linearly interpolated $B$ and $\bm x$ ($B_0=B_1$, $\bm x_0=0$) and $\|I-e^{-q_k/n}\mathbf R(\bm\omega_k/n)\|\le(q_k+\|\bm\omega_k\|)/n$, $M_F+V_F$ stays bounded independently of $n$, bounding the output difference by $K/n$ even when $\lambda$ and $\mu$ vary across substeps.

Write $\mathcal A_T$ and $\mathcal A$ for the right-hand sides of \eqref{eq:surrogate} and \eqref{eq:volterra}, $\mathcal B_T$ for $\mathcal A_T$ at $\lambda\equiv1$, and $\Pi_Tw=(w(s_1),\dots,w(s_T))$ under $\|\bm w\|_\infty=\max_i\|w_i\|$. Let $M_B,M_C,M_W$ bound $B,C,W$ on $[0,1]\times\Srms$ and set $M_v=M_BM_W$. Constants below depend only on these bounds, the Lipschitz constants in Theorem~\ref{thm:volterra}, $\|S_o\|$, $d$, $\mathrm{Lip}\,u_0$ and $t_f$.

\begin{proof}[Proof of Theorem~\ref{thm:volterra}]
\emph{State transitions.} Since $x\mapsto e^{-x}$ is 1-Lipschitz on $[0,\infty)$, rotations are isometries, and $\|\mathbf R(\bm a)-\mathbf R(\bm b)\|\le\|\bm a-\bm b\|$,
\begin{equation}\label{eq:transition-error}
\|U_{ij}-U(s_i,s_j)\|\le2\varepsilon_T,
\end{equation}
and $U(s,\sigma)$ and $\kappa$ are bounded and Lipschitz on their domains.

\emph{Well-posedness.} Extending $B$, $C$ and $W$ to bounded Lipschitz maps on $[0,1]\times\R^d$ (e.g., McShane extension composed with radial retraction) makes $\mathcal A$ locally Lipschitz on $C([0,1],\R^d)$, yielding a unique local solution. Writing $\mathcal J[u]$ for $S_o$ times the integral in \eqref{eq:volterra}, $\rho=\|u\|^2/d$ satisfies $\partial_t\rho=\frac2d(1-\rho)\langle u,\mathcal J[u]\rangle$ with $\rho(\cdot,0)=1$, so $\rho\equiv1$ and boundedness on $\Srms$ gives global existence on $[0,t_f]$ (and likewise for \eqref{eq:surrogate}, since $\|U_{ij}\|\le1$ and $\sum_k\delta_k=1$). Splitting the integral at $s<s'$ into $[0,s]$ and $[s,s']$ and using $\|P_a-P_b\|\le2\|a-b\|/\sqrt d$ gives $\|\mathcal A[w](s)-\mathcal A[w](s')\|\le K_1(|s-s'|+\|w(s)-w(s')\|)$ for sphere-valued $w$. Gr\"onwall's inequality on $g(t)=\|u(s,t)-u(s',t)\|$ yields $\mathrm{Lip}\,u(\cdot,t)\le(\mathrm{Lip}\,u_0+K_1t)e^{K_1t}$ for $t\le t_f$.

\emph{Consistency.} Evaluated on $u_k=u(s_k,t)$, the mesh outputs of $\mathcal A_T[\Pi_Tu]$ and $\mathcal B_T[\Pi_Tu]$ differ by $-\sum_{k\le i}\delta_k(1-\lambda_k)(F_{ik}-F_{i,k-1})$. Here $q_k+\|\bm\omega_k\|\le(L_H+L_\Phi)\delta_k+2\varepsilon_T$, $\|\bm v_k-\bm v_{k-1}\|\le C\delta_k$ by position regularity, and $\|I-\alpha_k\mathbf R_k\|\le q_k+\|\bm\omega_k\|$, so $\|F_{ik}-F_{i,k-1}\|\le K_2(\delta_k+\varepsilon_T)$ for $k\ge2$, while the $k=1$ term is bounded by $\delta_1\|F_{i1}\|$. Since $|1-\lambda_k|\le1$ and $\sum_k\delta_k=1$, $\|\mathcal A_T[\Pi_Tu]-\mathcal B_T[\Pi_Tu]\|_\infty=O(\delta_{\max}+\varepsilon_T)$ uniformly in $\lambda$. Next, \eqref{eq:transition-error} replaces $U_{ik}$ by $U(s_i,s_k)$ in $\mathcal B_T[\Pi_Tu]$ at cost $O(\varepsilon_T)$, leaving a right-endpoint Riemann sum for $\Pi_T\mathcal A[u]$ with error $O(\delta_{\max})$. Thus
\begin{equation}\label{eq:field-consistency}
\|\mathcal A_T[\Pi_Tu]-\Pi_T\mathcal A[u]\|_\infty\le K(\delta_{\max}+\varepsilon_T).
\end{equation}

\emph{Stability and convergence.} In $\bm y_i=\sum_{j\le i}w_{ij}F_{ij}$, the frozen weights $w_{ii}=\lambda_i\delta_i$ and $w_{ij}=\lambda_j\delta_j+(1-\lambda_{j+1})\delta_{j+1}$ ($j<i$) are nonnegative with row sum $\lambda_1\delta_1+\sum_{k=2}^i\delta_k\le1$. Boundedness and Lipschitz continuity of $B,C,W$ together with the tangent-projection bound yield $\|\mathcal A_T[\bm z]-\mathcal A_T[\bm w]\|_\infty\le L\|\bm z-\bm w\|_\infty$ on $\Srms^T$ uniformly in $T$ and $\lambda$. Combining this stability bound with \eqref{eq:field-consistency} gives $D^+e_T\le Le_T+K(\delta_{\max}+\varepsilon_T)$ for $e_T(t)=\|\bm u(t)-\Pi_Tu(\cdot,t)\|_\infty$, and Gr\"onwall's inequality completes the proof.
\end{proof}

Smoothness is essential even in the limit. Take $\delta_k=q_k=1/T$ with $\alpha=e^{-1/T}$, $C_i=1$, no rotation, $v_0=0$ and oscillating inputs $v_k=(-1)^k$. For $\lambda_k=(1+(-1)^k)/2$, \eqref{eq:difference} gives $y_T^\lambda-y_T^{1/2}=\frac{\alpha^{T-1}}{2T}+\frac{(1+\alpha)(1-\alpha^{T-1})}{2T(1-\alpha)}\to1-e^{-1}$. For constant $\lambda$ the sum telescopes to $(\lambda-1/2)(-1)^T/T\to0$. Finally, under Mamba-3's decay floor $q_k\ge A_{\rm floor}\Delta_k$ with $A_{\rm floor}>0$ (Appendix~\ref{app:setup}), appending context tokens onto a fixed prefix ($\Delta_1$ fixed) cannot realize the bounded-decay regime of Theorem~\ref{thm:volterra}: with $S_T=\sum_{k\le T}\Delta_k$, sending $\delta_{\max}\ge\Delta_1/S_T\to0$ forces $S_T\to\infty$ and hence $H_T(1)\ge A_{\rm floor}S_T\to\infty$.

\subsection{Numerical check}\label{app:numerical}

We tested Theorem~\ref{thm:volterra} numerically on 20 random smooth systems. Each system has $d=N=P=4$, two rotation planes and $S_o=I$. Its maps $B,C,W$ are SiLU maps with smooth spatial harmonics, and its decay and angular rates are smooth and positive. We integrated each system to $t_f=1.5$ with 120 projected RK4 steps. The meshes are uniform, smoothly warped or of alternating width, with $T=16,32,64,128,256$. We used six schedules of $\lambda$: $\lambda\equiv0,1/2,1$, a smooth harmonic schedule, an alternating $0/1$ schedule and a pseudo-random $0/1$ sequence. Two quantities shrink at an observed rate near $T^{-1}$. The first is the median worst-rule sup-norm error against a 4{,}096-point uniform right-endpoint reference, which falls from $7.1\text{--}8.0\times10^{-3}$ at $T=16$ to $3.6\text{--}4.8\times10^{-4}$ at $T=256$. The second is the median six-rule pairwise diameter, which falls from $1.1\text{--}1.2\times10^{-2}$ to $6.2\text{--}6.9\times10^{-4}$.

\subsection{A head with constant parameters}\label{app:constant}

Take constant $\Delta$, $q>0$ and $\lambda$, no rotation ($U_{ij}=\alpha^{i-j}I$ with $\alpha=e^{-q}$), and fixed $B,C$ with $B^\top C\ne0$. In \eqref{eq:quadrature}, the current-token weight is $k_0=\lambda\Delta$ and each past weight at lag $m\ge1$ is $\lambda\Delta+(1-\lambda)\Delta=\Delta$, giving $\bm y_i=(B^\top C)\sum_{m=0}^{i-1}k_m\bm x_{i-m}$ with $k_m=\Delta\alpha^m$ for $m\ge1$. Summing over $m\ge0$ gives steady-state gain $G_\lambda=\Delta(\lambda+1/(e^q-1))$ (a constant input starting at token 1 has finite-time gain $G_\lambda-\Delta\alpha^i/(1-\alpha)$ at token $i$), so $G_\lambda-G_\mu=\Delta(\lambda-\mu)$ and
\[
\frac{G_\lambda}{\Delta/q}=\lambda q+\frac{q}{e^q-1}=1+\Bigl(\lambda-\frac12\Bigr)q+\frac{q^2}{12}+O(q^4)\qquad(q\to0).
\]

\begin{proof}[Proof of Proposition~\ref{prop:lagone}]
Summing over $m\ge0$ with $z=\alpha^2\in(0,1)$, Cauchy--Schwarz gives $1/p_1=\lambda^2/z+1/(1-z)=\bigl(\lambda^2/z+1/(1-z)\bigr)\bigl(z+(1-z)\bigr)\ge(1+\lambda)^2$, with equality for $\lambda>0$ if and only if $z=\lambda/(1+\lambda)$. At $\lambda=0$, $p_1=1-z\to1$ as $q\to\infty$.
\end{proof}

Truncating to lags $0,\dots,M$ gives $p_{1,M}=k_1^2/\sum_{m=0}^{M}k_m^2\le p_1/(1-\alpha^{2M})$, where $(1-\alpha^{2M})^{-1}<1+10^{-6}$ for $M=16$ and $q\ge1/2$. When rotation or token-varying decays are allowed, the lag-one share can exceed $4/9$ at $\lambda=1/2$:
(i) with rotation ($N=2$, $B=e_1$, $C=e_2$, and per-step angle $\pi/2$), $B^\top C=0$ kills lag zero while $K_m=\Delta\alpha^m\sin(m\pi/2)$ gives $K_1^2/\sum_{m\ge0}K_m^2=1-\alpha^4>4/9$ whenever $q>\tfrac14\log\tfrac95$, and
(ii) without rotation, setting $q_i=q_{\rm low}$ for even $i$ and $q_i=q_{\rm reset}$ for odd $i$ with constant $\Delta$, $\lambda$, $B$ and $C$ sends even outputs to coefficients $\lambda\Delta$ and $\Delta$ at lags zero and one and odd outputs to $\lambda\Delta$ at lag zero as $q_{\rm low}\to0$ and $q_{\rm reset}\to\infty$, giving pooled lag-one share $1/(1+2\lambda^2)$ across adjacent outputs ($2/3$ at $\lambda=1/2$).

\section{Experiments on the released models}\label{app:exp}
\suppressfloats[t]

\subsection{Setup and evaluation splits}\label{app:protocol}

We evaluate the four official SISO checkpoints of \citet{lahoti2026mamba3} in BF16 as released. Their identifiers are \nolinkurl{state-spaces/mamba3-siso-187m}, \nolinkurl{-443m}, \nolinkurl{-893m} and \nolinkurl{-1.5b}.\footnote{Our standalone evaluation uses the OpenAI LAMBADA test split and greedy exact match on all tokens of the final word. It scores 34.7\%, 43.2\%, 46.9\% and 48.3\%, against 32.5\%, 40.2\%, 47.2\% and 49.4\% from the LM Evaluation Harness in Table~3 of \citet{lahoti2026mamba3}.} Each PG-19 window contains a beginning-of-sequence token plus 2{,}047 tokens starting at offset 4{,}096 \citep{rae2020compressive}. Figure~\ref{fig:headline} and Table~\ref{tab:native} use 32 test books (set 1). Table~\ref{tab:dial} uses 32 disjoint test books (set 2), and Figure~\ref{fig:dial} uses the first 16. Medians and 95\% bootstrap intervals on PG-19 are computed across books. Four-tuples of model values below always appear in the order 187M, 443M, 893M, 1.5B.

For 443M and 1.5B, we chose the selected layer (layer 3) and the early control layers (1--4) on PG-19 set 1 before evaluating set 2. For 187M and 893M, we fixed the selection on 8 validation books before using any test books. The selected layer maximizes the $\lambda=1/2$ validation loss increase, which gives layers 2 and 1. The early control layers are 1 to $L/6$. Exploratory summaries are marked $^\dagger$.

\subsection{Current-token and past-token interventions}\label{app:split}

\begin{table}[!htbp]
\centering
\caption{PG-19 loss increases of $\lambda=1/2$ in nats per token, summed over one-layer interventions, and the current-token share on PG-19 sets 1 and 2. Current token and Past token apply $\lambda=1/2$ to one part. Interaction is the full loss increase minus the sum of the two separate parts. Medians over 32 books. $^\dagger$Exploratory.}
\label{tab:native}
\small
\input{tables/tab_native.tex}
\end{table}

\begin{table}[!htbp]
\centering
\caption{LAMBADA baseline accuracy (\%) of the four released SISO checkpoints and percentage points lost under $\lambda=1/2$ in the PG-19-selected early layer or across all layers together. Current token and Past token apply only one part of \eqref{eq:presentpast}. Means over 5{,}153 documents.}
\label{tab:lambada}
\small
\input{tables/tab_lambada.tex}
\end{table}

\paragraph{Exact split.}
Section~\ref{sec:native} splits each reset into the current-token and past-token changes of \eqref{eq:presentpast}. Let $a_i=\Delta_i(\lambda_i^{\rm learned}-1/2)$. The current-token change is $\mathcal D_i=a_i(B_i^\top C_i)\bm x_i$, with biases included and the rotation canceling on the diagonal. The past-token change is $\mathcal O_i=\sum_{j<i}(a_j-a_{j+1})F_{ij}$. The two parts satisfy $\bm y_i^{1/2}=\bm y_i^{\rm learned}-\mathcal D_i-\mathcal O_i$. Changing only the current-token term gives $\bm y_i^{\rm learned}-\mathcal D_i$. Changing only the past-token terms gives $\bm y_i^{1/2}+\mathcal D_i=\bm y_i^{\rm learned}-\mathcal O_i$. To apply the past-token change, we call the released kernel with $\lambda=1$, $D=0$ and rescaled inputs $\bm x'_j=(\delta w_j/\Delta_j)\,\bm x_j$, where $\delta w_j=a_{j+1}-a_j$. We subtract the current-token term of this call and add the result to $\bm y_i^{\rm learned}$. An independent float64 implementation agrees with this split to $10^{-14}$. Subdivision with $n=8$ divides both parts by 8 to $10^{-13}$. With all rescaled inputs set to zero, the kernel route reproduces the released logits bit for bit.

\paragraph{Current-token share.}
The current-token share of a book is $1-\sum_\ell\Delta L^{\rm past}_\ell/\sum_\ell\Delta L_\ell$ over separate one-layer interventions (Table~\ref{tab:native}). Its 95\% intervals lie within 0.89--0.93 on set 1 and 0.88--0.93 on set 2. We also recompute the share with the same formula under head-mean replacement $\lambda\mapsto\bar\lambda$ on set 2, as a descriptive measure. The medians are 36.2\%, 42.2\%, 63.3\% and 37.6\%, with 95\% intervals from 31.4 to 39.7, 38.3 to 45.1, 51.4 to 71.2 and 31.7 to 46.2. In 1.5B, a slightly negative $\Delta L^{\rm past}$ gives some books a share above one, and we leave these shares unclipped.

\subsection{Head groups and controls}\label{app:groups}

\paragraph{Groups and cost ratios.}
Table~\ref{tab:dial} and Table~\ref{tab:heads} group heads on the 8 validation books to locate the current-token cost. Early layers are 0--3 in 187M and 0--6 in the others. Late layers are 8--11 in 187M and 17--23 in the others. In Table~\ref{tab:dial}, the mean-$\lambda$ cost ratio divides the summed one-layer cost of resetting each head to its mean $\bar\lambda$ by the summed cost of resetting it to $1/2$. Its 95\% intervals lie within 0.09--0.22. Recovery shares are relative to restoring the current-token term in every head.

\paragraph{Controls.}
Table~\ref{tab:dial} compares the current-token change with two controls of the same size. Both control costs are relative to the cost of the current-token change in the low-$\lambda$ heads of the early control layers (1--2 in 187M, 1--4 in the others). The post-projection norm of a change is its norm after the output projection $W_{\rm out}$. The random control averages eight Gaussian draws that match, at every token, the post-projection norm of the current-token change in the low-$\lambda$ heads. The skip control rescales $D\mapsto(1+\kappa_\ell)D$ in those heads. We calibrate $\kappa_\ell$ on validation books to match the pooled post-projection norm and sign. The 95\% intervals of the skip-control cost ratios lie within 2.2--4.1.

\paragraph{Cancellation ratio.}
The ratio $R$ of Table~\ref{tab:dial} measures how far the current-token term cancels the skip term. Let $g_i=\lambda_i\Delta_iB_i^\top C_i$ be the current-token coefficient at the learned $\lambda_i$, and let $w_i=\|\bm x_i\odot\mathrm{SiLU}(\bm z_i)\|^2$. The gated ratio is $R=\sum_i(D+g_i)^2w_i/\sum_i(D^2+g_i^2)w_i$. We compute it over short-horizon low-$\lambda$ heads on the first 16 books of set 2. Its upper 95\% bounds are 0.14--0.24. We also inspected the individual signs of $D$ and $g_i$ and Figure~\ref{fig:dial}$^\dagger$. On the 16 test books of Figure~\ref{fig:dial}, 33 of 36, 86 of 91, 134 of 143 and 184 of 196 validation-selected low-$\lambda$ heads lie in the region $|\bar\lambda-1/2|\,\bar q>0.1$$^\dagger$.

\begin{table}[!htbp]
\centering
\caption{Groups of heads for PG-19 set 2 and the median $\lambda$ of the short-horizon groups. Near $1/2$ is the share of long-horizon heads with $\bar\lambda$ within 0.1 of $1/2$.}
\label{tab:heads}
\small
\input{tables/tab_heads.tex}
\end{table}

\subsection{Other corpora}\label{app:xcorpus}

We repeat the layerwise evaluation on three other corpora (Table~\ref{tab:xcorpus}). They are WikiText-103 validation articles \citep{merity2017pointer}, Python files from \nolinkurl{codeparrot/codeparrot-clean-valid} (one per repository), and documents from the first shard of the 10B-token FineWeb-Edu sample \citep{penedo2024fineweb}. Each window holds a beginning-of-sequence token plus the first 2{,}047 tokens of a source. Each corpus has 8 verification windows and 32 evaluation windows. We keep the PG-19 validation-selected low-$\lambda$ heads without reselection. Past-token changes use the $\lambda=1$ kernel route of Appendix~\ref{app:split}, which avoids BF16 rounding in the final-layer subtraction and matches float64 within $\sim$1\%.

The 95\% interval of the layerwise current-token share lies above $0.5$ in all 12 model--corpus pairs. The interval of the low-$\lambda$ recovery minus $0.7$ times the all-head recovery lies above zero in 11 of 12 pairs. For Python in 893M, this interval is $[-0.04,0.07]$, with median recovery 0.72 (Table~\ref{tab:xcorpus}). Across heads, $\bar\lambda$ correlates with the PG-19 validation means at Spearman 0.98--0.99 on prose and 0.89--0.92 on Python.

\begin{table}[!htbp]
\centering
\caption{The current-token change on three other corpora, medians over 32 windows with the lower ends of their 95\% intervals. $\Delta L$ is summed over one-layer interventions, in nats per token.}
\label{tab:xcorpus}
\small
\input{tables/tab_xcorpus.tex}
\end{table}

\subsection{Setting low-\texorpdfstring{$\lambda$}{lambda} heads to zero}\label{app:zero-control}

Section~\ref{sec:native} compares the endpoints $\lambda=0$ and $\lambda=1/2$ in the low-$\lambda$ heads. This follow-up uses 32 unused PG-19 train books, selected from the first train shard by URL hash. Their overlap with the pretraining data of the released checkpoints is unknown. Each window holds a beginning-of-sequence token plus 2{,}047 tokens from offset 4{,}096. We intervene on the validation-selected low-$\lambda$ heads of the selected early layer (11, 9, 31 and 29 heads). Table~\ref{tab:zero-control} compares full replacement by $\lambda=0$ or $\lambda=1/2$, the separate parts of $\lambda=0$, and the current-token change of $\lambda=1/2$. Setting $\lambda=0$ costs 22 to 560 times less than $\lambda=1/2$. The exploratory per-book interaction is the full $\lambda=0$ increase minus the sum of its two parts. In 187M, it has median 0.01861 nats (interval from 0.01360 to 0.02116) and comes from downstream layers$^\dagger$. Float64 checks of the decomposition agree within $4\times10^{-12}$ on random inputs and within 0.8\%--1.2\% on 256-token post-projection layer changes. The direct and summed parts of the BF16 kernel agree within 0.3\%--0.5\%.

\begin{table}[!htbp]
\centering\small
\caption{Follow-up endpoint interventions in the selected low-$\lambda$ heads. Entries are median PG-19 loss increases in nats per token over 32 books. Lower and Upper are the ends of the 95\% bootstrap interval (20{,}000 resamples) of the full $\lambda=0$ increase.}
\label{tab:zero-control}
\setlength{\tabcolsep}{4pt}
\begin{tabular}{@{}lcrrrrrrr@{}}
\toprule
& & \multicolumn{5}{c}{Target $\lambda=0$} & \multicolumn{2}{c}{Target $\lambda=1/2$} \\
\cmidrule(lr){3-7}\cmidrule(l){8-9}
Model & Layer & Full & Lower & Upper & \twoline{Current}{token} & \twoline{Past}{token} & Full & \twoline{Current}{token} \\
\midrule
187M & 2 & 0.04266 & 0.03928 & 0.05352 & 0.02147 & 0.00469 & 0.95760 & 0.96968 \\
443M & 3 & 0.00089 & 0.00002 & 0.00249 & 0.00148 & 0.00049 & 0.49812 & 0.50557 \\
893M & 1 & 0.00242 & 0.00207 & 0.00413 & 0.00297 & 0.00063 & 0.30667 & 0.31485 \\
1.5B & 3 & 0.00173 & 0.00052 & 0.00269 & 0.00163 & 0.00057 & 0.23338 & 0.23649 \\
\bottomrule
\end{tabular}
\end{table}

\subsection{Continuum-assumption diagnostics}\label{app:assumptions}

\begin{figure}[!htbp]
\centering
\includegraphics[width=\linewidth]{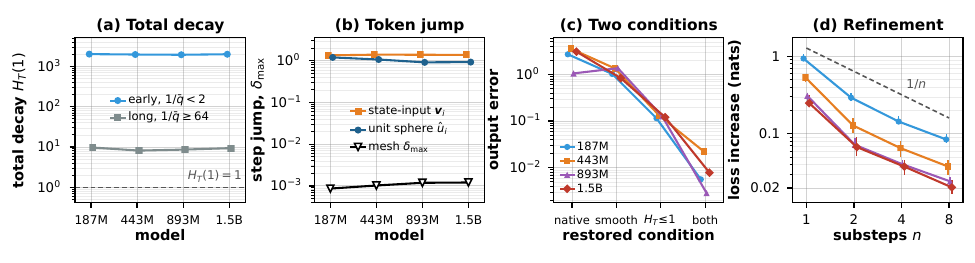}
\caption{Diagnostics of assumptions (A1) and (A2) of Theorem~\ref{thm:volterra} on the first 16 PG-19 test books. (a) Cumulative decay of long-horizon heads and of early short-horizon heads. (b) Relative jumps between adjacent tokens, against the normalized mesh width $\delta_{\max}$. (c) Relative error of the SSM output under $\lambda=1/2$ in the selected layer's low-$\lambda$ heads, on unmodified inputs, with smoothed inputs, with rescaled decay and rotation, and with both. (d) PG-19 loss increase of $\lambda=1/2$ over the learned rule when the selected layer is split into $n$ substeps under both rules. Medians with 95\% bootstrap intervals.}
\label{fig:assumptions}
\end{figure}

\paragraph{(A1) and (A2) in the checkpoints.}
Section~\ref{sec:native} checks assumptions (A1) and (A2) of Theorem~\ref{thm:volterra} in the released checkpoints. Figure~\ref{fig:assumptions} evaluates them on the first 16 PG-19 test books ($T=2{,}048$, steps $i\ge3$ for jumps)$^\dagger$. For 443M and 1.5B, the selected layer was chosen on PG-19 set 1, which contains these books. Let $q_{0.98}$ be the 98th percentile of $q_i$ over tokens. Early short-horizon heads have $\delta_{\max}=0.86\text{--}1.21\times10^{-3}$. They still reach $H_T(1)=1{,}940\text{--}2{,}013$ ($\bar q=0.95\text{--}0.98$, $q_{0.98}=2.10\text{--}2.31$) and total rotations $\sum_i\|\bm\omega_i\|=5{,}223\text{--}6{,}260$. Long-horizon heads reach $H_T(1)=8.1\text{--}9.7$ ($\bar q=0.0040\text{--}0.0047$, $q_{0.98}=0.014\text{--}0.020$) and total rotations of $94\text{--}126$. Let $\hat u_i=u_i/\|u_i\|$ be the unit-sphere layer input, where $u_i$ is the pre-normed input of the Mamba-3 layer (Appendix~\ref{app:setup}). In the selected layer, it jumps by $\|\hat u_i-\hat u_{i-1}\|=0.91\text{--}1.20$ between adjacent tokens. The relative state-input jumps $2\|\bm v_i-\bm v_{i-1}\|_F/(\|\bm v_i\|_F+\|\bm v_{i-1}\|_F)$ are $1.34\text{--}1.39$, or $2{,}667\text{--}2{,}866\,\delta_i$. The one-step increments $2\|F_{ii}-F_{i,i-1}\|/(\|F_{ii}\|+\|F_{i,i-1}\|)$ are $1.49\text{--}1.61$. Long-horizon heads have similar state-input jumps ($1.32\text{--}1.37$). The jumps therefore fail the smoothness condition in every head group, while the decay separates the two groups.

\paragraph{The $2\times2$ test.}
Section~\ref{sec:native} enforces each assumption in the selected early layer and compares the SSM outputs of the learned rule and $\lambda=1/2$. The test runs in float64 under normalized steps $\delta_i$. We rescale decay and rotation by multiplying $(q_i,\bm\omega_i)$ by $1/\max\{H_T(1),1\}$ in each head and book. We smooth the inputs by applying a norm-preserving Gaussian filter ($\sigma=64$ tokens) to $u_i$, with $(\Delta,A,\bm\omega,\lambda)$ fixed. The smoothed inputs change by $\|\Delta\hat u_i\|=0.9\text{--}2.3\times10^{-3}$. The test measures the relative error $\|\bm y^{\rm learned}-\bm y^{1/2}\|_F/\|\bm y^{\rm learned}\|_F$. Its median over low-$\lambda$ heads is $1.05\text{--}3.59$ on unmodified inputs and $0.83\text{--}1.36$ with smoothed inputs. Smoothing raises it in 893M, to 1.36 against 1.05. The median is $0.115\text{--}0.131$ with rescaled decay and rotation and $0.0029\text{--}0.0224$ with both. Of the 11, 9, 31 and 29 low-$\lambda$ heads, 8, 7, 26 and 21 are short-horizon. The relative error pooled over all heads of the layer is $0.14\text{--}0.24$ unmodified and $0.012\text{--}0.025$ with both. Smoothing alone already gives $0.012\text{--}0.025$ in three of the four models, and $0.57$ in 893M.

\paragraph{Subdivision.}
Figure~\ref{fig:assumptions}d checks the identity \eqref{eq:identity} by splitting the selected layer into $n$ linearly interpolated substeps. At $n=8$, the local $\lambda=1/2$ change shrinks by $7.997\text{--}7.999\times$, as the identity requires. The final-logit change shrinks by $5.47,7.50,6.73,7.36\times$, and the PG-19 $\Delta L$ falls from $0.25\text{--}0.94$ to $0.020\text{--}0.085$ nats. Without subdivision, $\lambda=1/2$ also shifts the decays of the downstream layers $\ell>\ell_{\rm sel}$. Let $q^{1/2}$ and $q^{\rm learned}$ be these decays under $\lambda=1/2$ and under the learned rule. The relative shift $|q^{1/2}-q^{\rm learned}|/\bar q^{\rm learned}$ has medians of $18\%\text{--}40\%$ when one layer is reset and $54\%\text{--}67\%$ when all layers are reset$^\dagger$.

\section{The training study}\label{app:training}

\subsection{Training and evaluation setup}\label{app:train-setup}

We train the 187M architecture under each rule of Section~\ref{sec:training} (A, D, H, B) for 5{,}578 steps on batches of $128\times2{,}048$ FineWeb-Edu tokens, 1.46B tokens in total. Training uses the Llama-3.1 tokenizer and AdamW with $(\beta_1,\beta_2)=(0.9,0.95)$. The learning rate peaks at $2\times10^{-3}$ after 300 warmup steps and follows a cosine decay to 10\%. Matrix weight decay is 0.1, the gradient clip is 1.0, and training uses BF16 autocast. Within each seed, the four rules share the initialization, the data order and the code path. We report seeds 4--6. Three earlier pilot seeds per rule served only to set up the tests. In H, the current-token coefficient $c_i\Delta_i$ equals $\Delta_i/2$ to BF16 precision. Checkpoints are saved at 0, 10M, 25M, 50M, 75M, 100M, 125M, 150M, 200M, 250M, 300M, 500M, 1B and 1.46B tokens.

Evaluation uses 46 unused PG-19 books in three disjoint sets, with 2{,}048-token windows from offset 4{,}096. We select the heads of layers 1--3 with final $1/\bar q<2$ on validation books 16--23, which gives 14--20 heads per run (Table~\ref{tab:training_runs}). We measure head statistics on validation books 24--31 and intervention losses on the remaining 30 books. They are test books 80--83, 85, 87, 88, 90--94 and validation books 32--49. Validation loss uses the first 1{,}024 sequences of a separate FineWeb-Edu token file (\texttt{val013}).

\subsection{Head measurements}\label{app:train-measure}

\paragraph{Alignment.}
Figure~\ref{fig:training}a reports the alignment of the $B$ and $C$ projections. On the measurement books, a head's alignment is its mean signed cosine between $B_i$ and $C_i$, with biases included. $G$ is the run median of this alignment. We also record each head's mean $|\cos(B_i,C_i)|$. Across seeds 4--6, A and D have run-median $\lambda$ of $0.01\text{--}0.04$ and $G$ of $0.40\text{--}0.74$, within $0.001$ of the median $|\cos(B_i,C_i)|$. B and H have median $|\cos(B_i,C_i)|$ of $0.05\text{--}0.09$ and median $G$ between $-0.06$ and $0.08$.

\paragraph{Lag shares.}
The lag shares of Figure~\ref{fig:training}b come from the coefficients of the SSM output under rule $(\ell,c)$. The coefficient is $c_i\Delta_iB_i^\top C_i$ at lag zero and $e^{-(H_i-H_{i-m})}\eta_{i-m}^\ell C_i^\top\mathbf R(\bm\Phi_i-\bm\Phi_{i-m})B_{i-m}$ at lag $m\ge1$, where $\eta_j^\ell=\ell_j\Delta_j+(1-\ell_{j+1})\Delta_{j+1}$. We normalize the squared coefficients of each head by their sum over lags 0--16. The median lag-one share is $0.77\text{--}0.88$ across all runs and $0.77\text{--}0.85$ in B and H. The lag-zero share is $0.005\text{--}0.020$ in A and D and $0.034\text{--}0.093$ in B and H. Replacing lag zero by $D+c_i\Delta_iB_i^\top C_i$ in the same denominator adds the skip term. It gives median lag-zero shares of $0.011\text{--}0.012$ (A), $0.006\text{--}0.009$ (D), $0.042\text{--}0.048$ (H) and $0.031\text{--}0.041$ (B) across seeds 4--6. The same replacement lowers the lag-zero share of the 15 short-horizon heads in layers 1--3 of the released 187M checkpoint from 0.175 to 0.049. The gated current-token term carries 0.5\%--0.8\% of squared gated head output in A and D and 1.8\%--2.4\% in H and B.

\subsection{Tests}\label{app:train-tests}

Tables~\ref{tab:training_summary} and~\ref{tab:training_runs} report two tests. The lag-one test subtracts the lag-one contribution of the selected heads, one layer at a time, and compares the loss increase with that of a random vector of the same post-projection norm. The swap test changes one factor of $c_i\Delta_iB_i^\top C_i$ at a time in layers 1--3 and keeps the state, the past-token terms, the skip term and the gate fixed. It pairs A with H and D with B. The small-factor swap sets $\lambda\mapsto1/2$ in A and D and $B_i^\top C_i\mapsto G\|B_i\|\|C_i\|$ in H and B, with $G$ from the paired run. The other-factor swap sets the other factor to its value in the paired run. The all-layer current-token share is the fraction of the all-layer $\lambda=1/2$ loss increase that restoring the learned current-token term in all layers removes, as a median over books. In rule A at the end of training, it is 0.930, 0.922 and 0.940 in seeds 4--6.

\begin{table}[!htbp]
\centering\small
\caption{The 187M training runs by seed. Heads counts the short-horizon heads in layers 1--3. Loss gaps are paired means relative to rule A over 1{,}024 FineWeb-Edu validation sequences. The swap and lag-one tests give book medians of summed PG-19 loss increases in nats per token.}
\label{tab:training_runs}
\input{tables/tab_training_final.tex}
\end{table}

\subsection{Training dynamics}\label{app:dynamics}

\paragraph{Short horizon before small $\lambda$.}
Section~\ref{sec:training} reports that most heads in A and D reach a short horizon before a small $\lambda$. The count covers all 96 heads of layers 0--3 (Figure~\ref{fig:evolution}e). A head counts as short-horizon or low-$\lambda$ once its condition holds at two consecutive checkpoints. Among heads that become both, the short horizon comes first in 17/21, 14/19 and 12/20 heads under A and in 20/24, 19/23 and 15/21 under D ($60\%\text{--}83\%$). In total, 49, 44 and 41 heads become low-$\lambda$ under A, against 79, 77 and 76 under D.

\paragraph{Trajectories.}
Figure~\ref{fig:evolution}a--d,f,g tracks the alignment, lag-one share and $\lambda$ of the selected heads and the reset cost of rule A through training. The selected heads start at $|\cos(B_i,C_i)|=0.49\text{--}0.50$ and lag-one share $0.14\text{--}0.17$. In A and D, $|\cos(B_i,C_i)|$ dips to $0.19\text{--}0.35$ at 50M--125M tokens and rises again from 75M--150M. In these two rules, $\lambda$ first drops below $0.1$ between 75M and 200M tokens. Rule A's summed one-layer $\lambda=1/2$ cost grows from $0.03\text{--}0.06$ nats at 25M to $1.09\text{--}2.28$ nats at 1.46B. Its current-token share grows from $7\%\text{--}22\%$ to $88\%\text{--}92\%$ over the same span. The all-layer current-token share is evaluated at 100M, 250M, 500M and 1B tokens and at the end of training.

\subsection{Paired validation losses}\label{app:paired}

The loss gaps in Table~\ref{tab:training_runs} are paired means relative to rule A. We also paired the validation losses of B with those of H and D over the 1{,}024 validation sequences within each seed. The loss of B exceeds that of H by $0.0065$, $0.0035$ and $0.0072$ nats in seeds 4, 5 and 6, with 95\% bootstrap intervals from $0.0054$ to $0.0075$, from $0.0024$ to $0.0046$ and from $0.0062$ to $0.0083$. It exceeds that of D by $0.0057$, $0.0006$ and $0.0029$ nats, with intervals from $0.0047$ to $0.0068$, from $-0.0004$ to $0.0017$ and from $0.0019$ to $0.0040$. These intervals resample the validation sequences within one pair of runs and do not measure the variation between training runs.

\end{document}

%% file: tables/tab_dial.tex
\begin{tabular}{@{}lcccccccc@{}}
\toprule
& \multicolumn{2}{c}{Selected layer $\Delta L$} & \multicolumn{3}{c}{Summed over layers} & & \multicolumn{2}{c}{Control ratio} \\
\cmidrule(lr){2-3}\cmidrule(lr){4-6}\cmidrule(l){8-9}
Model & $\lambda=1/2$ & Restored & \twoline{Mean-$\lambda$}{cost ratio} & \twoline{Low-$\lambda$}{recovery} & \twoline{Other}{recovery} & $R$ & Random & Skip \\
\midrule
187M & 0.99 & 0.022 & 0.10 & 0.91 & 0.09 & 0.13 & 0.35 & 3.6\\
443M & 0.52 & 0.007 & 0.12 & 0.90 & 0.09 & 0.18 & 0.25 & 2.3\\
893M & 0.31 & 0.007 & 0.19 & 0.84 & 0.16 & 0.20 & 0.19 & 2.7\\
1.5B & 0.26 & 0.007 & 0.12 & 0.88 & 0.11 & 0.23 & 0.24 & 3.5\\
\bottomrule
\end{tabular}

%% file: tables/tab_training_summary.tex
\begin{tabular}{@{}lcccccccc@{}}
\toprule
& \multicolumn{2}{c}{Rule weight} & & \multicolumn{3}{c}{Swap test} & \multicolumn{2}{c}{Lag-one test} \\
\cmidrule(lr){2-3}\cmidrule(lr){5-7}\cmidrule(l){8-9}
Rule & Current & Past & Loss gap & \twoline{Small}{factor} & Random & \twoline{Other}{factor} & Lag one & Random \\
\midrule
A & learned & learned & -- & 0.70 & 0.27 & \phantom{$-$}0.000 & 0.095 & 0.037 \\
D & learned & $1/2$ & 0.003 & 0.73 & 0.29 & $-$0.001 & 0.105 & 0.042 \\
H & $1/2$ & learned & 0.003 & 0.25 & 0.10 & \phantom{$-$}0.000 & 0.115 & 0.037 \\
B & $1/2$ & $1/2$ & 0.009 & 0.36 & 0.15 & \phantom{$-$}0.001 & 0.133 & 0.045 \\
\bottomrule
\end{tabular}

%% file: tables/tab_native.tex
\begin{tabular}{@{}lccccccccc@{}}
\toprule
& \multicolumn{4}{c}{Summed $\Delta L$ (nats per token)} & \multicolumn{5}{c}{Current-token share} \\
\cmidrule(lr){2-5}\cmidrule(l){6-10}
Model & $\lambda=1/2$ & \twoline{Current}{token} & \twoline{Past}{token} & Interaction & \twoline{Set}{1} & \twoline{Set}{2} & \twoline{Other}{layers$^\dagger$} & \twoline{Late}{half$^\dagger$} & Joint$^\dagger$ \\
\midrule
187M & 1.68 & 1.51 & 0.13 & 0.013 & 0.92 & 0.92 & 0.82 & 0.71 & 0.95 \\
443M & 1.19 & 1.04 & 0.11 & 0.023 & 0.91 & 0.91 & 0.84 & 0.66 & 0.97 \\
893M & 0.96 & 0.84 & 0.09 & 0.018 & 0.91 & 0.91 & 0.87 & 0.75 & 0.97 \\
1.5B & 0.72 & 0.65 & 0.06 & 0.020 & 0.91 & 0.92 & 0.88 & 0.74 & 0.97 \\
\bottomrule
\end{tabular}

%% file: tables/tab_lambada.tex
\begin{tabular}{@{}lcccccccc@{}}
\toprule
& & \multicolumn{4}{c}{Selected layer} & \multicolumn{3}{c}{All layers} \\
\cmidrule(lr){3-6}\cmidrule(l){7-9}
Model & Accuracy & Layer & $\lambda=1/2$ & \twoline{Current}{token} & \twoline{Past}{token} & $\lambda=1/2$ & \twoline{Current}{token} & \twoline{Past}{token} \\
\midrule
187M & 34.7 & 2 & 20.3 & 19.9 & 0.7 & 34.6 & 31.6 & 15.0 \\
443M & 43.2 & 3 & 18.8 & 18.9 & 0.3 & 43.0 & 42.6 & 14.0 \\
893M & 46.9 & 1 & \phantom{0}8.6 & \phantom{0}8.8 & 0.4 & 46.8 & 46.2 & 12.9 \\
1.5B & 48.3 & 3 & 11.6 & 11.7 & 0.3 & 48.2 & 48.0 & 13.3 \\
\bottomrule
\end{tabular}

%% file: tables/tab_heads.tex
\begin{tabular}{@{}lccccccccc@{}}
\toprule
& & & & \multicolumn{2}{c}{Short, early} & \multicolumn{2}{c}{Short, late} & \multicolumn{2}{c}{Long} \\
\cmidrule(lr){5-6}\cmidrule(lr){7-8}\cmidrule(l){9-10}
Model & All & Low $\lambda$ & \twoline{Short,}{low $\lambda$} & Heads & $\lambda$ & Heads & $\lambda$ & Heads & Near $1/2$ \\
\midrule
187M & \phantom{0}288 & \phantom{0}36 & \phantom{0}28 & \phantom{0}37 & 0.14 & \phantom{0}40 & 0.66 & \phantom{0}80 & 0.79 \\
443M & \phantom{0}768 & \phantom{0}91 & \phantom{0}77 & \phantom{0}85 & 0.11 & \phantom{0}95 & 0.60 & 219 & 0.75 \\
893M & 1152 & 143 & 119 & 121 & 0.11 & 127 & 0.53 & 327 & 0.75 \\
1.5B & 1536 & 196 & 155 & 145 & 0.10 & 158 & 0.56 & 464 & 0.81 \\
\bottomrule
\end{tabular}

%% file: tables/tab_xcorpus.tex
\begin{tabular}{@{}llccccc@{}}
\toprule
& & & \multicolumn{2}{c}{Current-token share} & \multicolumn{2}{c}{Recovery share, low $\lambda$} \\
\cmidrule(lr){4-5}\cmidrule(l){6-7}
Corpus & Model & $\Delta L$ & Median & Lower end & Median & Lower end \\
\midrule
WikiText-103 & 187M & 2.21 & 0.92 & 0.91 & 0.94 & 0.94 \\
 & 443M & 1.27 & 0.89 & 0.88 & 0.94 & 0.94 \\
 & 893M & 0.73 & 0.86 & 0.84 & 0.88 & 0.88 \\
 & 1.5B & 0.73 & 0.84 & 0.83 & 0.92 & 0.91 \\
\addlinespace[2pt]
Python & 187M & 4.42 & 0.90 & 0.89 & 0.93 & 0.92 \\
 & 443M & 2.97 & 0.84 & 0.84 & 0.89 & 0.88 \\
 & 893M & 2.58 & 0.88 & 0.87 & 0.72 & 0.68 \\
 & 1.5B & 2.22 & 0.89 & 0.87 & 0.87 & 0.86 \\
\addlinespace[2pt]
FineWeb-Edu & 187M & 1.68 & 0.94 & 0.93 & 0.93 & 0.92 \\
 & 443M & 0.89 & 0.92 & 0.90 & 0.92 & 0.92 \\
 & 893M & 0.59 & 0.91 & 0.88 & 0.89 & 0.88 \\
 & 1.5B & 0.58 & 0.89 & 0.86 & 0.91 & 0.90 \\
\bottomrule
\end{tabular}

%% file: tables/tab_training_final.tex
\begin{tabular}{@{}llccccccc@{}}
\toprule
& & & & \multicolumn{3}{c}{Swap test} & \multicolumn{2}{c}{Lag-one test} \\
\cmidrule(lr){5-7}\cmidrule(l){8-9}
Rule & Seed & Heads & Loss gap & \twoline{Small}{factor} & Random & \twoline{Other}{factor} & Lag one & Random \\
\midrule
A & 4 & 14 & -- & 0.41 & 0.17 & \phantom{$-$}0.000 & 0.079 & 0.030 \\
 & 5 & 17 & -- & 0.70 & 0.27 & \phantom{$-$}0.000 & 0.095 & 0.037 \\
 & 6 & 17 & -- & 1.52 & 0.69 & $-$0.002 & 0.104 & 0.039 \\
\addlinespace[2pt]
D & 4 & 17 & 0.003 & 0.44 & 0.16 & \phantom{$-$}0.000 & 0.110 & 0.042 \\
 & 5 & 18 & 0.002 & 0.73 & 0.29 & $-$0.001 & 0.102 & 0.044 \\
 & 6 & 17 & 0.008 & 0.99 & 0.35 & $-$0.003 & 0.105 & 0.040 \\
\addlinespace[2pt]
H & 4 & 15 & 0.003 & 0.12 & 0.05 & \phantom{$-$}0.002 & 0.115 & 0.037 \\
 & 5 & 20 & 0.000 & 0.25 & 0.10 & \phantom{$-$}0.000 & 0.133 & 0.042 \\
 & 6 & 16 & 0.004 & 0.43 & 0.13 & \phantom{$-$}0.000 & 0.111 & 0.037 \\
\addlinespace[2pt]
B & 4 & 18 & 0.009 & 0.21 & 0.08 & \phantom{$-$}0.000 & 0.154 & 0.048 \\
 & 5 & 18 & 0.003 & 0.36 & 0.15 & \phantom{$-$}0.003 & 0.133 & 0.045 \\
 & 6 & 16 & 0.011 & 0.38 & 0.17 & \phantom{$-$}0.001 & 0.121 & 0.043 \\
\bottomrule
\end{tabular}

%% file: refs.bib
@inproceedings{chen2018neuralode,
  author    = {Chen, Ricky T. Q. and Rubanova, Yulia and Bettencourt, Jesse and Duvenaud, David K.},
  title     = {Neural Ordinary Differential Equations},
  booktitle = {Advances in Neural Information Processing Systems (NeurIPS)},
  volume    = {31},
  year      = {2018},
  url       = {https://proceedings.neurips.cc/paper/2018/hash/69386f6bb1dfed68692a24c8686939b9-Abstract.html},
}

@inproceedings{kidger2020neuralcde,
  author    = {Kidger, Patrick and Morrill, James and Foster, James and Lyons, Terry},
  title     = {Neural Controlled Differential Equations for Irregular Time Series},
  booktitle = {Advances in Neural Information Processing Systems (NeurIPS)},
  volume    = {33},
  pages     = {6696--6707},
  year      = {2020},
  url       = {https://proceedings.neurips.cc/paper/2020/hash/4a5876b450b45371f6cfe5047ac8cd45-Abstract.html},
}

@inproceedings{geshkovski2023emergence,
  author    = {Geshkovski, Borjan and Letrouit, Cyril and Polyanskiy, Yury and Rigollet, Philippe},
  title     = {The Emergence of Clusters in Self-Attention Dynamics},
  booktitle = {Advances in Neural Information Processing Systems (NeurIPS)},
  volume    = {36},
  pages     = {57026--57037},
  year      = {2023},
  url       = {https://proceedings.neurips.cc/paper_files/paper/2023/hash/b2b3e1d9840eba17ad9bbf073e009afe-Abstract-Conference.html},
}

@inproceedings{karagodin2024clustering,
  author    = {Karagodin, Nikita and Polyanskiy, Yury and Rigollet, Philippe},
  title     = {Clustering in Causal Attention Masking},
  booktitle = {Advances in Neural Information Processing Systems (NeurIPS)},
  volume    = {37},
  pages     = {115652--115681},
  year      = {2024},
  doi       = {10.52202/079017-3673},
  url       = {https://proceedings.neurips.cc/paper_files/paper/2024/hash/d18d208fa9c333483e5724ade7beff0f-Abstract-Conference.html},
}

@article{rigollet2025meanfield,
  author  = {Rigollet, Philippe},
  title   = {The Mean-Field Dynamics of Transformers},
  journal = {arXiv preprint arXiv:2512.01868},
  year    = {2025},
  url     = {https://arxiv.org/abs/2512.01868},
}

@inproceedings{gu2020hippo,
  author    = {Gu, Albert and Dao, Tri and Ermon, Stefano and Rudra, Atri and R{\'e}, Christopher},
  title     = {{HiPPO}: Recurrent Memory with Optimal Polynomial Projections},
  booktitle = {Advances in Neural Information Processing Systems (NeurIPS)},
  volume    = {33},
  pages     = {1474--1487},
  year      = {2020},
  url       = {https://proceedings.neurips.cc/paper_files/paper/2020/hash/102f0bb6efb3a6128a3c750dd16729be-Abstract.html},
}

@inproceedings{gu2021lssl,
  author    = {Gu, Albert and Johnson, Isys and Goel, Karan and Saab, Khaled and Dao, Tri and Rudra, Atri and R{\'e}, Christopher},
  title     = {Combining Recurrent, Convolutional, and Continuous-time Models with Linear State Space Layers},
  booktitle = {Advances in Neural Information Processing Systems (NeurIPS)},
  volume    = {34},
  pages     = {572--585},
  year      = {2021},
  url       = {https://proceedings.neurips.cc/paper/2021/hash/05546b0e38ab9175cd905eebcc6ebb76-Abstract.html},
}

@inproceedings{gu2023mamba,
  author    = {Gu, Albert and Dao, Tri},
  title     = {Mamba: Linear-Time Sequence Modeling with Selective State Spaces},
  booktitle = {Conference on Language Modeling (COLM)},
  year      = {2024},
  url       = {https://openreview.net/forum?id=tEYskw1VY2},
}

@inproceedings{gu2022s4,
  author    = {Gu, Albert and Goel, Karan and R{\'e}, Christopher},
  title     = {Efficiently Modeling Long Sequences with Structured State Spaces},
  booktitle = {International Conference on Learning Representations (ICLR)},
  year      = {2022},
  url       = {https://openreview.net/forum?id=uYLFoz1vlAC},
}

@inproceedings{dao2024mamba2,
  author    = {Dao, Tri and Gu, Albert},
  title     = {Transformers are {SSMs}: Generalized Models and Efficient Algorithms Through Structured State Space Duality},
  booktitle = {Proceedings of the 41st International Conference on Machine Learning},
  series    = {Proceedings of Machine Learning Research},
  volume    = {235},
  pages     = {10041--10071},
  year      = {2024},
  publisher = {PMLR},
  url       = {https://proceedings.mlr.press/v235/dao24a.html},
}

@inproceedings{lahoti2026mamba3,
  author    = {Lahoti, Aakash and Li, Kevin Y. and Chen, Berlin and Wang, Caitlin and Bick, Aviv and Kolter, J. Zico and Dao, Tri and Gu, Albert},
  title     = {Mamba-3: Improved Sequence Modeling using State Space Principles},
  booktitle = {International Conference on Learning Representations (ICLR)},
  year      = {2026},
  url       = {https://openreview.net/forum?id=HwCvaJOiCj},
}

@inproceedings{shen2026upi,
  author    = {Shen, Haochen and Wertheimer, Davis and Wang, Zheng and Goon, Garrett and Liu, Derrick and Wang, Naigang and Srivatsa, Mudhakar and Ganti, Raghu K. and Zhang, Minjia},
  title     = {From Collapse to Control: Understanding and Extending Context Length in Emerging Hybrid Models via {Universal Position Interpolation}},
  booktitle = {International Conference on Learning Representations (ICLR)},
  year      = {2026},
  url       = {https://openreview.net/forum?id=MjmORKLHUI},
}

@article{hochbruck2010exponential,
  author  = {Hochbruck, Marlis and Ostermann, Alexander},
  title   = {Exponential Integrators},
  journal = {Acta Numerica},
  volume  = {19},
  pages   = {209--286},
  year    = {2010},
  doi     = {10.1017/S0962492910000048},
}

@inproceedings{vo2024demystifying,
  author    = {Vo, Thieu and Pham, Duy-Tung and Tong, Xin T. and Nguyen, Tan Minh},
  title     = {Demystifying the Token Dynamics of Deep Selective State Space Models},
  booktitle = {International Conference on Learning Representations (ICLR)},
  pages     = {47349--47375},
  year      = {2025},
  url       = {https://proceedings.iclr.cc/paper_files/paper/2025/hash/756dfc9113ae4f310847530403fcb212-Abstract-Conference.html},
}

@article{nguyen2025tokenspheres,
  author    = {Nguyen, Trinh Tien and Nguyen-Nhat, Minh-Khoi and Pham, Duy-Tung and Do, Hoang-Son and Nguyen, Tan Minh and Vo, Thieu},
  title     = {Token Dynamics on Spheres in {Mamba} Models},
  journal   = {OpenReview preprint},
  year      = {2025},
  url       = {https://openreview.net/forum?id=466gVY2sBQ},
}

@inproceedings{cirone2024theoretical,
  author    = {Muca Cirone, Nicola and Orvieto, Antonio and Walker, Benjamin and Salvi, Cristopher and Lyons, Terry},
  title     = {Theoretical Foundations of Deep Selective State-Space Models},
  booktitle = {Advances in Neural Information Processing Systems (NeurIPS)},
  volume    = {37},
  pages     = {127226--127272},
  year      = {2024},
  publisher = {Curran Associates, Inc.},
  doi       = {10.52202/079017-4041},
  url       = {https://proceedings.neurips.cc/paper_files/paper/2024/hash/e6231c5f46598cfd09ff1970524e0436-Abstract-Conference.html},
}

@article{castin2025unified,
  author  = {Castin, Val\'{e}rie and Ablin, Pierre and Carrillo, Jos\'{e} Antonio and Peyr\'{e}, Gabriel},
  title   = {A Unified Perspective on the Dynamics of Deep Transformers},
  journal = {arXiv preprint arXiv:2501.18322},
  year    = {2025},
  url     = {https://arxiv.org/abs/2501.18322},
}

@inproceedings{endy2025mambaknockout,
  author    = {Endy, Nir and Grosbard, Idan Daniel and Ran-Milo, Yuval and Slutzky, Yonatan and Tshuva, Itay and Giryes, Raja},
  title     = {{Mamba} Knockout for Unraveling Factual Information Flow},
  booktitle = {Proceedings of the 63rd Annual Meeting of the Association for Computational Linguistics (Volume 1: Long Papers)},
  pages     = {23457--23477},
  year      = {2025},
  publisher = {Association for Computational Linguistics},
  doi       = {10.18653/v1/2025.acl-long.1143},
  url       = {https://aclanthology.org/2025.acl-long.1143/},
}

@inproceedings{pitorro2025latim,
  author    = {Pitorro, Hugo and Treviso, Marcos Vinicius},
  title     = {{LaTIM}: Measuring Latent Token-to-Token Interactions in {Mamba} Models},
  booktitle = {Proceedings of the 63rd Annual Meeting of the Association for Computational Linguistics (Volume 1: Long Papers)},
  pages     = {24478--24493},
  year      = {2025},
  publisher = {Association for Computational Linguistics},
  doi       = {10.18653/v1/2025.acl-long.1194},
  url       = {https://aclanthology.org/2025.acl-long.1194/},
}

@inproceedings{hossain2025selectivememory,
  author    = {Hossain, Tamanna and {Logan IV}, Robert L. and Jagadeesan, Chandrasekhara Ganesh and Singh, Sameer and Tetreault, Joel R. and Jaimes, Alejandro},
  title     = {Characterizing {Mamba}'s Selective Memory using Auto-Encoders},
  booktitle = {Proceedings of the 14th International Joint Conference on Natural Language Processing and the 4th Conference of the Asia-Pacific Chapter of the Association for Computational Linguistics},
  pages     = {2007--2022},
  year      = {2025},
  publisher = {The Asian Federation of Natural Language Processing and The Association for Computational Linguistics},
  doi       = {10.18653/v1/2025.ijcnlp-long.109},
  url       = {https://aclanthology.org/2025.ijcnlp-long.109/},
}

@book{gripenberg1990volterra,
  author    = {Gripenberg, G. and Londen, S. O. and Staffans, O.},
  title     = {Volterra Integral and Functional Equations},
  series    = {Encyclopedia of Mathematics and its Applications},
  volume    = {34},
  publisher = {Cambridge University Press},
  address   = {Cambridge},
  year      = {1990},
  doi       = {10.1017/CBO9780511662805},
  url       = {https://doi.org/10.1017/CBO9780511662805},
}

@inproceedings{lu2025mambamodulation,
  author    = {Lu, Peng and Huang, Jerry and Zeng, Qiuhao and Wang, Xinyu and Chen, Boxing and Langlais, Philippe and Cui, Yufei},
  title     = {Mamba Modulation: On the Length Generalization of {Mamba} Models},
  booktitle = {Advances in Neural Information Processing Systems (NeurIPS)},
  volume    = {38},
  pages     = {19414--19459},
  year      = {2025},
  publisher = {Curran Associates, Inc.},
  url       = {https://proceedings.neurips.cc/paper_files/paper/2025/hash/1bfc9f74afa91b9b8add5a97a97001a1-Abstract-Conference.html},
}

@inproceedings{benkish2024decimamba,
  author    = {Ben-Kish, Assaf and Zimerman, Itamar and Abu-Hussein, Shady and Cohen, Nadav and Globerson, Amir and Wolf, Lior and Giryes, Raja},
  title     = {{DeciMamba}: Exploring the Length Extrapolation Potential of {Mamba}},
  booktitle = {International Conference on Learning Representations (ICLR)},
  pages     = {101148--101170},
  year      = {2025},
  url       = {https://proceedings.iclr.cc/paper_files/paper/2025/hash/fae0a4535196bf7715c1aef2ccfe283f-Abstract-Conference.html},
}

@article{zappala2023neuralide,
  author    = {Zappala, Emanuele and de Oliveira Fonseca, Antonio Henrique and Moberly, Andrew Henry and Higley, Michael James and Abdallah, Chadi and Cardin, Jessica A. and van Dijk, David},
  title     = {Neural Integro-Differential Equations},
  journal   = {Proceedings of the AAAI Conference on Artificial Intelligence},
  volume    = {37},
  number    = {9},
  pages     = {11104--11112},
  year      = {2023},
  doi       = {10.1609/aaai.v37i9.26315},
  url       = {https://ojs.aaai.org/index.php/AAAI/article/view/26315},
}

@inproceedings{rae2020compressive,
  author    = {Rae, Jack W. and Potapenko, Anna and Jayakumar, Siddhant M. and Hillier, Chloe and Lillicrap, Timothy P.},
  title     = {Compressive Transformers for Long-Range Sequence Modelling},
  booktitle = {International Conference on Learning Representations (ICLR)},
  year      = {2020},
  url       = {https://openreview.net/forum?id=SylKikSYDH},
}

@inproceedings{paperno2016lambada,
  author    = {Paperno, Denis and Kruszewski, Germ{\'a}n and Lazaridou, Angeliki and Pham, Ngoc Quan and Bernardi, Raffaella and Pezzelle, Sandro and Baroni, Marco and Boleda, Gemma and Fern{\'a}ndez, Raquel},
  title     = {The {LAMBADA} Dataset: Word Prediction Requiring a Broad Discourse Context},
  booktitle = {Proceedings of the 54th Annual Meeting of the Association for Computational Linguistics (Volume 1: Long Papers)},
  pages     = {1525--1534},
  year      = {2016},
  publisher = {Association for Computational Linguistics},
  doi       = {10.18653/v1/P16-1144},
  url       = {https://aclanthology.org/P16-1144/},
}


%% file: refs_native.bib
@inproceedings{peng2023rwkv,
  title     = {{RWKV}: Reinventing {RNN}s for the {Transformer} Era},
  author    = {Peng, Bo and Alcaide, Eric and Anthony, Quentin and Albalak, Alon and Arcadinho, Samuel and Biderman, Stella and Cao, Huanqi and Cheng, Xin and Chung, Michael and Derczynski, Leon and Du, Xingjian and Grella, Matteo and Gv, Kranthi and He, Xuzheng and Hou, Haowen and Kazienko, Przemyslaw and Kocon, Jan and Kong, Jiaming and Koptyra, Bart{\l}omiej and Lau, Hayden and Lin, Jiaju and Mantri, Krishna Sri Ipsit and Mom, Ferdinand and Saito, Atsushi and Song, Guangyu and Tang, Xiangru and Wind, Johan and Wo{\'z}niak, Stanis{\l}aw and Zhang, Zhenyuan and Zhou, Qinghua and Zhu, Jian and Zhu, Rui-Jie},
  booktitle = {Findings of the Association for Computational Linguistics: EMNLP 2023},
  pages     = {14048--14077},
  publisher = {Association for Computational Linguistics},
  year      = {2023},
  doi       = {10.18653/v1/2023.findings-emnlp.936},
  url       = {https://aclanthology.org/2023.findings-emnlp.936/},
}

@inproceedings{merity2017pointer,
  author    = {Merity, Stephen and Xiong, Caiming and Bradbury, James and Socher, Richard},
  title     = {Pointer Sentinel Mixture Models},
  booktitle = {International Conference on Learning Representations (ICLR)},
  year      = {2017},
  url       = {https://openreview.net/forum?id=Byj72udxe},
}

@inproceedings{penedo2024fineweb,
  author    = {Penedo, Guilherme and Kydl{\'i}{\v{c}}ek, Hynek and Ben Allal, Loubna and Lozhkov, Anton and Mitchell, Margaret and Raffel, Colin and Von Werra, Leandro and Wolf, Thomas},
  title     = {The {FineWeb} Datasets: Decanting the Web for the Finest Text Data at Scale},
  booktitle = {Advances in Neural Information Processing Systems (NeurIPS), Datasets and Benchmarks Track},
  volume    = {37},
  pages     = {30811--30849},
  year      = {2024},
  url       = {https://proceedings.neurips.cc/paper_files/paper/2024/hash/370df50ccfdf8bde18f8f9c2d9151bda-Abstract-Datasets_and_Benchmarks_Track.html},
}

@inproceedings{wang2025universality,
  author    = {Wang, Junxuan and Ge, Xuyang and Shu, Wentao and Tang, Qiong and Zhou, Yunhua and He, Zhengfu and Qiu, Xipeng},
  title     = {Towards Universality: Studying Mechanistic Similarity Across Language Model Architectures},
  booktitle = {International Conference on Learning Representations (ICLR)},
  year      = {2025},
  url       = {https://openreview.net/forum?id=2J18i8T0oI},
}

@article{arora2025mechanistic,
  author  = {Arora, Aryaman and Rathi, Neil and Selvam, Nikil Roashan and Csord{\'a}s, R{\'o}bert and Jurafsky, Dan and Potts, Christopher},
  title   = {Mechanistic Evaluation of {Transformers} and State Space Models},
  journal = {arXiv preprint arXiv:2505.15105},
  year    = {2025},
  url     = {https://arxiv.org/abs/2505.15105},
}

@inproceedings{fu2023h3,
  author    = {Fu, Daniel Y. and Dao, Tri and Saab, Khaled K. and Thomas, Armin W. and Rudra, Atri and R{\'e}, Christopher},
  title     = {Hungry Hungry Hippos: Towards Language Modeling with State Space Models},
  booktitle = {International Conference on Learning Representations (ICLR)},
  year      = {2023},
  url       = {https://openreview.net/forum?id=COZDy0WYGg},
}

@article{elhage2021framework,
  author  = {Elhage, Nelson and Nanda, Neel and Olsson, Catherine and Henighan, Tom and Joseph, Nicholas and Mann, Ben and Askell, Amanda and Bai, Yuntao and Chen, Anna and Conerly, Tom and DasSarma, Nova and Drain, Dawn and Ganguli, Deep and Hatfield-Dodds, Zac and Hernandez, Danny and Jones, Andy and Kernion, Jackson and Lovitt, Liane and Ndousse, Kamal and Amodei, Dario and Brown, Tom and Clark, Jack and Kaplan, Jared and McCandlish, Sam and Olah, Chris},
  title   = {A Mathematical Framework for {Transformer} Circuits},
  journal = {Transformer Circuits Thread},
  year    = {2021},
  url     = {https://transformer-circuits.pub/2021/framework/index.html},
}

@inproceedings{wang2023ioi,
  author    = {Wang, Kevin Ro and Variengien, Alexandre and Conmy, Arthur and Shlegeris, Buck and Steinhardt, Jacob},
  title     = {Interpretability in the Wild: a Circuit for Indirect Object Identification in {GPT-2} Small},
  booktitle = {International Conference on Learning Representations (ICLR)},
  year      = {2023},
  url       = {https://arxiv.org/abs/2211.00593},
}

@article{mcgrath2023hydra,
  author  = {McGrath, Thomas and Rahtz, Matthew and Kram{\'a}r, J{\'a}nos and Mikulik, Vladimir and Legg, Shane},
  title   = {The {Hydra} Effect: Emergent Self-repair in Language Model Computations},
  journal = {arXiv preprint arXiv:2307.15771},
  year    = {2023},
  url     = {https://arxiv.org/abs/2307.15771},
}

@inproceedings{rushing2024selfrepair,
  author    = {Rushing, Cody and Nanda, Neel},
  title     = {Explorations of Self-Repair in Language Models},
  booktitle = {International Conference on Machine Learning (ICML)},
  series    = {Proceedings of Machine Learning Research},
  volume    = {235},
  pages     = {42836--42855},
  year      = {2024},
  url       = {https://proceedings.mlr.press/v235/rushing24a.html},
}

@article{heimersheim2024layernorm,
  author  = {Heimersheim, Stefan},
  title   = {You Can Remove {GPT2}'s {LayerNorm} by Fine-tuning},
  journal = {arXiv preprint arXiv:2409.13710},
  year    = {2024},
  url     = {https://arxiv.org/abs/2409.13710},
}

@inproceedings{baroni2026layernorm,
  author    = {Baroni, Luca and Khara, Galvin and Schaeffer, Joachim and Subkhankulov, Marat and Heimersheim, Stefan},
  title     = {Small Transformers Don't Need {LayerNorm} at Inference Time: Scaling {LayerNorm} Removal to {GPT-2 XL} and Implications for Mechanistic Interpretability},
  booktitle = {International Conference on Learning Representations (ICLR)},
  year      = {2026},
  url       = {https://openreview.net/forum?id=VPtHqcafIY},
}
